\documentclass{article} 
\usepackage{iclr2020_conference,times}

\usepackage{amsmath,amsfonts,bm}

\def\eqref#1{equation~\ref{#1}}

\def\1{\bm{1}}

\DeclareMathAlphabet{\mathsfit}{\encodingdefault}{\sfdefault}{m}{sl}
\SetMathAlphabet{\mathsfit}{bold}{\encodingdefault}{\sfdefault}{bx}{n}

\DeclareMathOperator*{\argmin}{arg\,min}

\usepackage{hyperref}
\usepackage{url}

\usepackage{booktabs}       
\usepackage{amsfonts}       
\usepackage{nicefrac}       
\usepackage{microtype}      

\usepackage{graphicx}
\usepackage{subcaption}
\usepackage{dsfont}
\usepackage{todonotes}
\usepackage{wrapfig}
\usepackage{listings}
\usepackage{verbatim}
\usepackage[normalem]{ulem}
\useunder{\uline}{\ul}{}
\usepackage{amsthm}

\theoremstyle{theorem}
\newtheorem{proposition}{Proposition}

\title{Detecting Underspecification with Local \\ Ensembles}

\author{%
  David Madras\thanks{Some of this work was done while this author was at Google Brain.} \\
  University of Toronto\\
  Vector Institute\\
  \texttt{madras@cs.toronto.edu} \\
   \And
   James Atwood \\
   Google Brain \\
   \texttt{atwoodj@google.com}
   \And 
   Alex D'Amour \\
   Google Brain \\
   \texttt{alexdamour@google.com}
}

\iclrfinalcopy 
\begin{document}

\maketitle

\begin{abstract}
We present \textit{local ensembles}, a method for detecting underspecification
--- when many possible predictors are consistent with the training data and model class --- 
at test time in a pre-trained model.
Our method uses local second-order information to approximate the variance of predictions across an ensemble of models from the same class. We compute this approximation by estimating the norm of the component of a test point's gradient that aligns with the low-curvature directions of the Hessian, and provide a tractable method for estimating this quantity. Experimentally, we show that our method is capable of detecting when a pre-trained model is underspecified on test data, with applications to out-of-distribution detection, detecting spurious correlates, and active learning.
\end{abstract}

\section{Introduction} \label{sec:intro}

As machine learning is deployed in increasingly vital areas, there is increasing demand for metrics that draw attention to potentially unreliable predictions.
One important source of unreliability is \textit{underspecification}.
Specifically, we say that a trained model is underspecified at a test input 
if many different predictions at that input are all equally consistent with the constraints posed by the training data and the learning problem specification (i.e., the model architecture and the loss function).
Equivalently, we can say that the prediction at that input is \textit{underdetermined}.
Underspecification is particularly relevant in the context of overparameterized model classes (e.g. deep neural networks). 
Recently, simple (but computationally expensive) ensembling methods \citep{lakshminarayanan2017simple}, which train many models on the same data from different random seeds, have proven highly effective at uncertainty quantification tasks \citep{ovadia2019can}.
This suggests that underspecification is a key threat to reliability in deep learning, and motivates flexible methods that can detect underdetermined predictions cheaply.

With this motivation, we present \textit{local ensembles}, a post-hoc method for measuring the extent to which a pre-trained model's prediction is underspecified for a particular test input.
Given a trained model, our method returns an underspecification score that measures the variability of test predictions across a \textit{local ensemble}, i.e. a set of local perturbations of the trained model parameters that fit the training data equally well.
Local ensembles are a computationally cheap, post-hoc alternative to fully trained ensembles and approximate Bayesian ensembling methods that require special training procedures \citep{gal2015dropout,blundell2015weight}.
Local ensembles also address a gap in approximate methods for estimating 
prediction uncertainty.
Specifically, whereas exact Bayesian or Frequentist uncertainty includes underspecification as one component, approximate methods such as Laplace approximations \citep{mackay1992information} or influence function-based methods \citep{schulam2019auditing} break down when underspecification is present. 
In contrast, our method leverages the pathology (an ill-conditioned Hessian) that makes these methods struggle.

Our contributions in this paper are as follows:

\begin{itemize}
    \item We identify underspecification as a key factor in the unreliability of predictions from overparametrized models, and present \textit{local ensembles}, a test-time method for detecting underspecification.
    \item We demonstrate theoretically that our method approximates the variance of a trained ensemble with local second-order information.
    \item We give a practical method for tractably approximating this quantity, which is simpler and cheaper than alternative second-order reliability methods.
    \item Through a set of experiments aimed at testing underspecification, we show our method approximates the behavior of trained ensembles, and can detect underspecification in a range of scenarios.
\end{itemize}{}
\section{Underspecification Score and Local Ensembles} \label{sec:lanczos}

\subsection{Setup}
Let $z = (x, y)$ be an example input-output pair, where $x$ is a vector of features and $y$ is a label.
We define a model in terms of a loss function $\mathcal L$ with parameters $\theta$ as a sum over training examples $(z_i)_{i=1}^n$, i.e., $\mathcal{L}(\theta) = \sum_i^n \ell(z_i, \theta)$, where $\ell$ is an example-wise loss (e.g., mean-squared error or cross entropy).
Let $\theta^{\star}$ be the parameters of the trained model, obtained by, e.g., minimizing the loss over this dataset, i.e., $\theta^{\star} = \argmin_{\theta} \mathcal{L}(\theta)$.
We write the prediction function given by parameters $\theta$ at an input $x$ as $\hat y(x, \theta)$.
We consider the problem of auditing a trained model, where unlabeled test points $x'$ arrive one at a time in a stream, and we wish to assess underspecification on a point-by-point basis.

\begin{wrapfigure}{R}{0.35\textwidth}
\centering
        \includegraphics[scale=0.33]{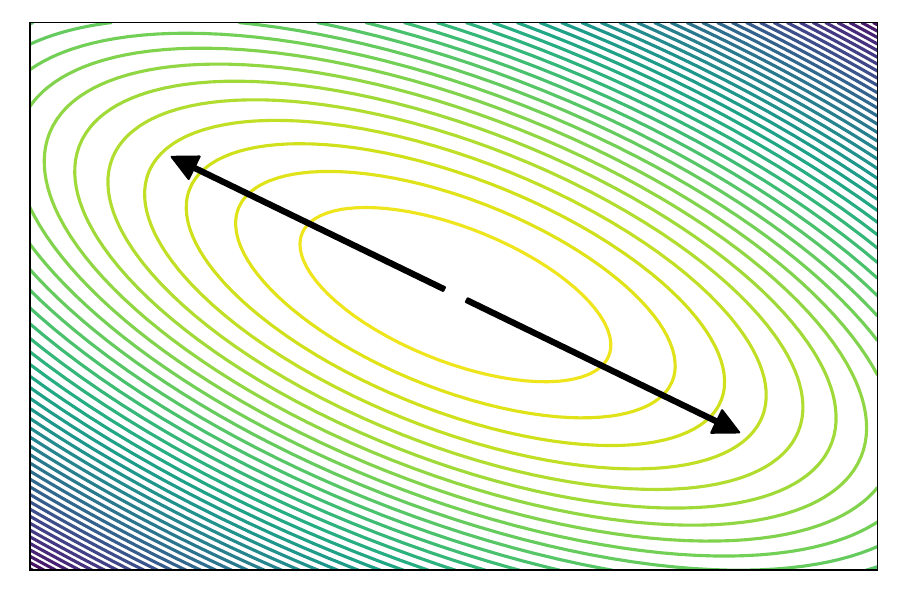}
        \caption{
            In this quadratic bowl, arrows denote the small eigendirection, where predictions are slow to change.
            We argue this direction is key to underspecification.
            }
        \label{fig:descriptive-quadratic-bowl}
\end{wrapfigure}
In this section, we introduce our local ensemble underspecification score $\mathcal E_m(x')$ for an unlabeled test point $x'$ (significance of $m$ explained below).
The score is designed to measure the variability that would be induced by randomly choosing predictions from an ensemble of models with similar training loss.
Our score has a simple form: it is the norm of the prediction gradient $g_{\theta^\star}(x') := \nabla_\theta \hat y(x', \theta^\star)$ multiplied by a matrix of Hessian eigenvectors spanning a subspace of low curvature $U_m$ (defined in more detail below).
\begin{equation}\label{eq:extrapolation score}
   \mathcal E_m(x') = \|U_m^\top g_{\theta^\star}(x')\|_2 
\end{equation}
Here, we show that this score is proportional to the standard deviation of predictions across a local ensemble of models with near-identical training loss, and demonstrate that this approximation holds in practice.

\subsection{Derivation}
Our derivation proceeds in two steps.
First, we define a local ensemble of models with similar training loss, then we state the relationship between our underspecification score and the variability of predictions within this ensemble.

The spectral decomposition of the Hessian $H_{\theta^\star}$ plays a key role in our derivation. Let
\begin{equation}
    H_{\theta^\star} = U \Lambda U^\top,
\end{equation}
where $U$ is a square matrix whose columns are the orthonormal eigenvectors of $H_{\theta^\star}$, written $(\xi_{(1)}, \cdots, \xi_{(p)})$, and $\Lambda$ is a square, diagonal matrix with the eigenvalues of $H_{\theta^\star}$, written $(\lambda_{(1)}, \cdots, \lambda_{(p)})$, along its diagonal.
As a convention, we index the eigenvectors and eigenvalues in decreasing order of the eigenvalue magnitude.

To construct a local ensemble of loss-preserving models, we exploit the fact that eigenvectors with large corresponding eigenvalues represent directions of high curvature, whereas eigenvectors with small corresponding eigenvalues represent directions of low curvature.
Thus, under the assumption that the model has been trained to a local minimum or saddle point, parameter perturbations in flat directions (those corresponding to small eigenvalues $\lambda_{(j)}$) do not change the training loss substantially (see Fig. \ref{fig:descriptive-quadratic-bowl}).
We characterize this subspace by the span of eigenvectors with corresponding small eigenvalues.
Formally, 
let $m$  be the eigenvalue index such that the eigenvalues $\{\lambda_{(j)} :  j > m\},$ are sufficiently small to be considered ``flat''.
\footnote{For now, we take $m$ to be given, and discuss tradeoffs for choosing $m$ in practice in Section~\ref{sec:computing-extrapolation-scores}.}
We call the subspace spanned by $\{\xi_{(j)} : j \in \sigma\}$ the \textit{ensemble subspace}.
Parameter perturbations in the ensemble subspace generate an ensemble of models with near-identical training loss.
\footnote{We note that not all loss-preserving ensembles will necessarily be local; however, the existence of some such local ensemble is sufficient to enable underdetermination.}

Our goal is to characterize the variance of test predictions with respect to random parameter perturbations $\Delta_\theta$ that occur in the ensemble subspace.
We now show that our underspecification score $\mathcal E_m(x')$ is, to the first order, proportional to the standard deviation of predictions at a point $x'$.

\begin{proposition}
Let $\Delta_\theta$ be the projection of a random perturbation with mean zero and covariance proportional to the identity $\epsilon \cdot I$ into the ensemble subspace spanned by $\{\xi_{(j)} : j > m\}$.
Let $P_\Delta$ be the linearized change in prediction induced by the perturbation
\begin{align*}
P_\Delta(x') &:=  g_{\theta^\star}(x')^\top \Delta_\theta \approx \hat y(x', \theta^\star + \Delta_\theta) - \hat y(x', \theta^\star).
\end{align*}

Then $\mathcal E_m(x') = \epsilon^{-1/2} \cdot SD(P_\Delta(x'))$.

\begin{proof}
Let $U_m$ be the matrix whose columns are $\{\xi_{(j)} : j > m\}$.
Then $U_m U_m^\top$ is a projection matrix that projects vectors into the ensemble subspace, and
$\Delta_\theta$ has covariance $\epsilon \cdot U_m U_m^\top$.
It follows that
\begin{align*}
\epsilon^{-1} Var(P_\Delta)
= g_{\theta^\star}(x')^\top U_m U_m^\top  g_{\theta^\star}(x')
= \|U_m^\top g_{\theta^\star}(x')\|^2_2  = \mathcal E_m(x')^2
\end{align*}
\end{proof}
\end{proposition}
\begin{wrapfigure}{L}{0.35\textwidth}
  \centering
  \includegraphics[scale=0.33]{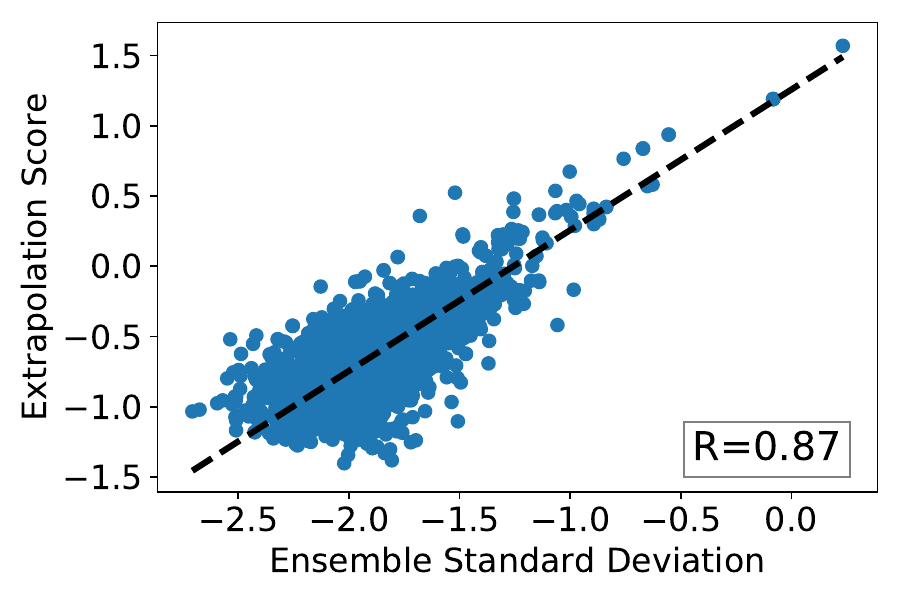}
  \caption{Mean underspecification score vs. ensemble prediction standard deviation on WineQuality dataset.
  Shows line of best fit and Pearson coefficient R.
  Both axes log-scaled.
  }
  \label{fig:ensembles}
\end{wrapfigure}

We test this hypothesized relationship on several tabular datasets.
We train an ensemble of twenty neural networks using the same architecture and training set, varying only the random seed.
Then, for each model in the ensemble, we calculate our local ensemble underspecification score $\mathcal{E}_m(x)$ for each input $x'$ in the test set (see Sec. \ref{sec:computing-extrapolation-scores} for details).
For each $x'$, we compare, across the ensemble, the mean value of $\mathcal{E}_m(x')$ to the standard deviation $\hat{y}(x')$.
In Fig. \ref{fig:ensembles}, we plot these two quantities against each other for one of the datasets, finding a nearly linear relationship.
On each dataset, we found a similar, significantly linear relationship (see Table \ref{table:ensembles-uci} and Appendix \ref{app:other-ensembles}).
We note the relationship is weaker with the Diabetes dataset; the standard deviations of these ensembles are an order of magnitude higher than the other datasets, indicating much noisier data.

Finally, we note that we can obtain similar results for the standard deviation of the loss at test points if we redefine $g$ as the loss gradient rather than the prediction gradient.
\begin{wraptable}{R}{0.3\textwidth}
    \centering
    \begin{tabular}{cc}
    \hline
    \textbf{Dataset} & \textbf{Pearson}  \\ \hline
    Boston           & 0.76             \\
    Diabetes         & 0.50            \\
    Abalone          & 0.76          \\
    Wine             & 0.87     \\ \hline
    \end{tabular}
	\caption{Correlation of underspecification scores and ensemble std. deviations on 4 datasets.}
	\label{table:ensembles-uci}
\end{wraptable}

\section{Computing Underspecification Scores} \label{sec:computing-extrapolation-scores}

We now discuss a practical method for computing our underspecification scores.
The key operation is constructing the set of eigenvectors that span a suitably loss-preserving ensemble subspace (i.e. have sufficiently small corresponding eigenvalues).
Because eigenvectors corresponding to large eigenvalues are easier to estimate, we construct this subspace by finding the top $m$ eigenvectors and defining the ensemble subspace as their orthogonal complement.
Our method can be implemented with any algorithm that returns the top $m$ eigenvectors.
See below for discussion on some tradeoffs in the choice of $m$, as well as the algorithm we choose (the Lanczos iteration \citep{lanczos1950iteration}).

Our method proceeds as follows.
For a given choice of $m$, we calculate the $m$ eigenvectors of $H$ with the \textit{largest} eigenvalues.
These eigenvectors define an $m$-dimensional subspace which is the orthogonal complement to the ensemble subspace.
We use these eigenvectors to construct a matrix that projects gradients into the ensemble subspace.
Specifically, let $U_{m^\perp}$ be the matrix whose columns are these large-eigenvalue eigenvectors $\{\xi_{(j)}: j \leq m\}$.
Then $U_{m^\perp} U_{m^\perp}^\top$ is the projection matrix that projects vectors into the ``large eigenvalue'' subspace, and $I - U_{m^\perp} U_{m^\perp}^\top$ projects into its complement: the ensemble subspace.
Now, for any test input $x'$, we take the norm of this projected gradient to compute our score 
$$
\mathcal E_m(x') = \left\| \left(I - U_{m^\perp}U_{m^\perp}^\top\right) g_{\theta^\star}(x') \right\|.
$$

The success of this approach depends on the eigenspectrum of the Hessian and the choice of $m$.
Specifically, the underspecification score $\mathcal E_m(x')$ is the most sensitive to underdetermination in the region of the trained parameters if we set $m$ to be the smallest index for which the training loss is relatively flat in the implied ensemble subspace.
If $m$ is set too low, the ensemble subspace will include well-constrained directions, and $\mathcal E_m(x')$ will over-estimate the prediction's sensitivity to loss-preserving perturbations.
If $m$ is set too high, the ensemble subspace will omit some under-constrained directions, and $\mathcal E_m(x')$ will be less sensitive.
For models where all parameters are well-constrained by the training data, a suitable $m$ may not exist.
This will usually not be the case for deep neural network models, which are known to have very ill-conditioned Hessians \citep[see, e.g., ][]{sagun2017empirical}.
It may be possible to develop diagnostics for choosing $m$ based on trace estimation techniques \citep{hutchinson1990stochastic} or eigenvalue distributions \citep{martin2018implicit,marvcenko1967distribution,wigner1967random}, but we leave this as future work.

We use the Lanczos iteration to estimate the top $m$ eigenvectors, which presents a number of practical advantages for usage in our scenario.
Firstly, it performs well under early stopping, returning good estimates of the top $m$ eigenvectors after $m$ iterations.
Secondly, we can cache intermediate steps, meaning that computing the $m + 1$-th eigenvector is fast once we have computed the first $m$.
Thirdly, it requires only implicit access to the Hessian through a function which applies matrix multiplication, meaning we can take advantage of efficient Hessian-vector product methods \citep{pearlmutter1994fast}.

Finally, the Lanczos iteration is simple -- it can be implemented in less than 20 lines of Python code (see Appendix \ref{app:lanczos-short-code}).
It contains only one hyperparameter, the stopping value $m$.
Fortunately, tuning this parameter is efficient --- given a maximum value $M$, we can try many values $m < M$ at once, by estimating $M$ eigenvectors and then calculating $\mathcal{E}_m$ by using the first $m$ eigenvectors.
The main constraint of our method is space rather than time --- while estimating the first $m$ eigenvectors enables easy caching for later use, it may be difficult to work with these eigenvectors in memory as $m$ and model size $p$ increase.
This tradeoff informed our choice of $m$ in this paper; we note in some cases that increasing $m$ further could have improved performance (see Appendix \ref{app:latent-factors}).
This suggests that further work on techniques for mitigating this tradeoff, e.g. online learning of sparse representations \citep{wang2016online,wang2012online}, could improve the performance of our method.
See Appendix \ref{app:lanczos} for more details on the Lanczos iteration.

\section{Related Work}

\subsection{Relation to Bayesian and Frequentist Second-Order Methods}\label{sec:comparison}
It is instructive to compare our underspecification score to two other approximate reliability quantification methods that are aimed at Bayesian and Frequentist notions of underspecification, respectively.
Like our underspecification score, both of these methods make use of local information in the Hessian to make an inference about the variance of a prediction.
First, consider the Laplace approximation of the posterior predictive variance.
This metric is derived by interpreting the loss function as being equivalent to a Bayesian log-posterior distribution over the model parameters $\theta$, and approximating it with a Gaussian.
Specifically, \citep[see, e.g.,][]{mackay1992information}
\begin{align}
Var(y \mid x') \approx g_{\theta^\star}(x')^\top H_{\theta^\star}^{-1} g_{\theta^\star} = \sum_{j=1}^p \lambda_{(j)}^{-1} \left(\xi_{(j)}^\top g_{\theta^\star}\right)^2.
\end{align}
Second, consider scores such as RUE (Resampling Under Uncertainty) designed to approximate the variability of predictions by resampling the training data \citep{schulam2019auditing}.
These methods approximate the change in trained parameter values induced by perturbing the training data via influence functions \citep{koh2017understanding,cook1982residuals}.
Specifically, the gradient of the parameters with respect to the weight of a given training example $z_i$ is given by 
\begin{align}
I(z_i) = -H_{\theta^\star}^{-1} \nabla_\theta \ell(z_i, \theta^\star) = \sum_{j=1}^p \lambda_{(j)}^{-2} \left(\xi_{(j)}^\top \nabla_\theta \ell(z_i, \theta^\star)\right)^2.
\end{align}
\citet{schulam2019auditing} combine this influence function with a specific random distribution of weights to approximate the variance of predictions under bootstrap resampling; other similar formulations are possible, sometimes with theoretical guarantees \citep{giordano2018swiss}.

Importantly, both of these methods work well when model parameters are well-constained by the training data, but they struggle when predictions are (close to) underdetermined.
This is because, in the presence of underdetermination, the Hessian becomes ill-conditioned.
Practical advice for dealing with this ill-conditioning is available \citep{koh2017understanding}, but we note that this not merely a numerical pathology; by our argument above, a poorly conditioned Hessian is a clear signal of underspecification.
In contrast to these methods, our method focuses specifically on prediction variability induced by underconstrained parameters.
Our underspecification score incorporates \emph{only} those terms with small eigenvalues, and removes the inverse eigenvalue weights that make inverse-Hessian methods break down.
This is clear from its summation representation: $\mathcal E_m(x') = \sum_{j > m} \left(\xi_{(j)}^\top g_{\theta^\star}\right)^2$.

Our method also has computational advantages over approaches that rely on inverting the Hessian.
Firstly, implicitly inverting the Hessian is a complex and costly process --- by finding only the important components of the projection explicitly, our method is simpler and more efficient.
Furthermore, we only need to find these components once; we can reuse them for future test inputs.
This type of caching is not possible with methods which require us to calculate the inverse Hessian-vector product for each new test input.

\subsection{Related Work in Detecting Underspecification}

Some recent works explore the relationship between test points, the learned model, and the training set.
Several papers examine reliability criteria that are based on distance in some space: within/between-group distances \citep{jiang2018trust}, a pre-specified kernel in a learned embedding space \citep{card2019deep}, or the activation space of a neural network \citep{papernot2018deep}.
We implement some nearest-neighbor baselines inspired by this work in Sec.~\ref{sec:experiments}.
Additionally, a range of methods exist for related tasks, such as OOD detection \citep{choi2018generative,liang2017enhancing,gal2015dropout,scholkopf2001estimating} and  calibration \citep{naeini2015obtaining,guo2017calibration}.
Some work using generative models for OOD detection makes use of related second-order analysis \citep{nalisnick2018deep}.
A line of work explores the benefits of training ensemble methods explicitly, discussed in detail in \citet{dietterich2000ensemble}.
These methods have been discussed for usage in some of the applications we present in Sec. \ref{sec:experiments}, including uncertainty detection \citep{lakshminarayanan2017simple}, active learning \citep{melville2004diverse} and OOD detection \citep{choi2018generative}.
Finally, a line of work on ``Rashomon sets'' explores loss-preserving ensembles in simpler model classes more formally \citep{semenova2019study,fisher2018all}.

\section{Experiments} \label{sec:experiments}

In this section, we give evidence that local ensembles can detect underspecification 
in trained models.
In order to explicitly evaluate underspecification, we present a range of experiments where a pre-trained model has a ``blind spot'', and evaluate its ability to detect when an input is in that blind spot.
We probe our method's ability to detect a range of underspecification, exploring cases where the blind spot is: \textbf{1.} easily visualized, \textbf{2.} well-defined by the feature distribution, \textbf{3.} well-defined by a latent distribution, and \textbf{4.} unknown, but where we can evaluate our model's detection performance through an auxiliary task.
See Appendix \ref{app:data-and-models} for experimental details.
Code for running the local ensembles method can be found at \url{https://github.com/dmadras/local-ensembles}.

\subsection{Visualizing Underspecification Detection} \label{sec:experiments-toy}

\begin{figure}[ht!]
  \begin{subfigure}[b]{.33\textwidth}
    \centering
        \includegraphics[scale=0.28]{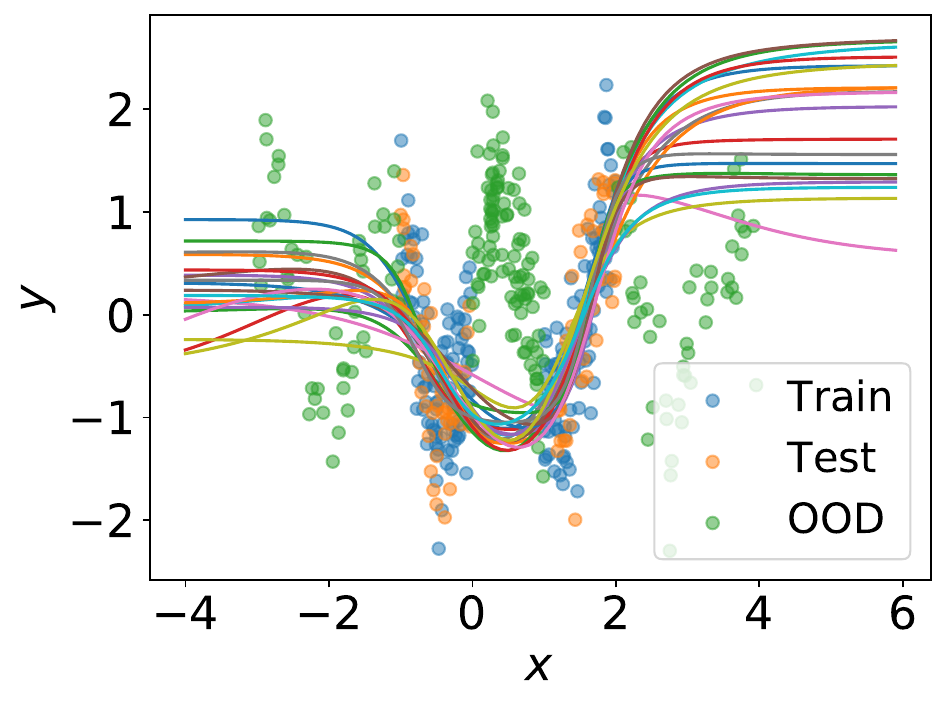}
        \caption{
            Data and trained model
            }
        \label{fig:small-NN-data}
  \end{subfigure}
  \begin{subfigure}[b]{.33\textwidth}
    \centering
        \includegraphics[scale=0.28]{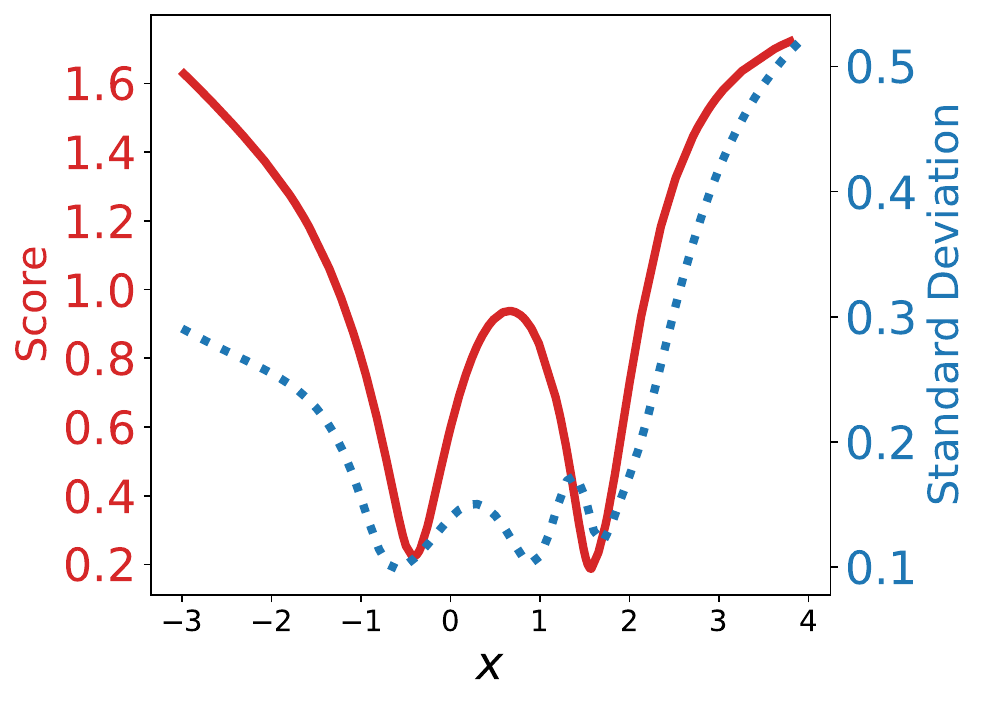}
        \caption{
        Underspecification scores
        }
        \label{fig:small-NN-scores-vs-stdev}
  \end{subfigure}
  \begin{subfigure}[b]{.33\textwidth}
    \centering
        \includegraphics[scale=0.28]{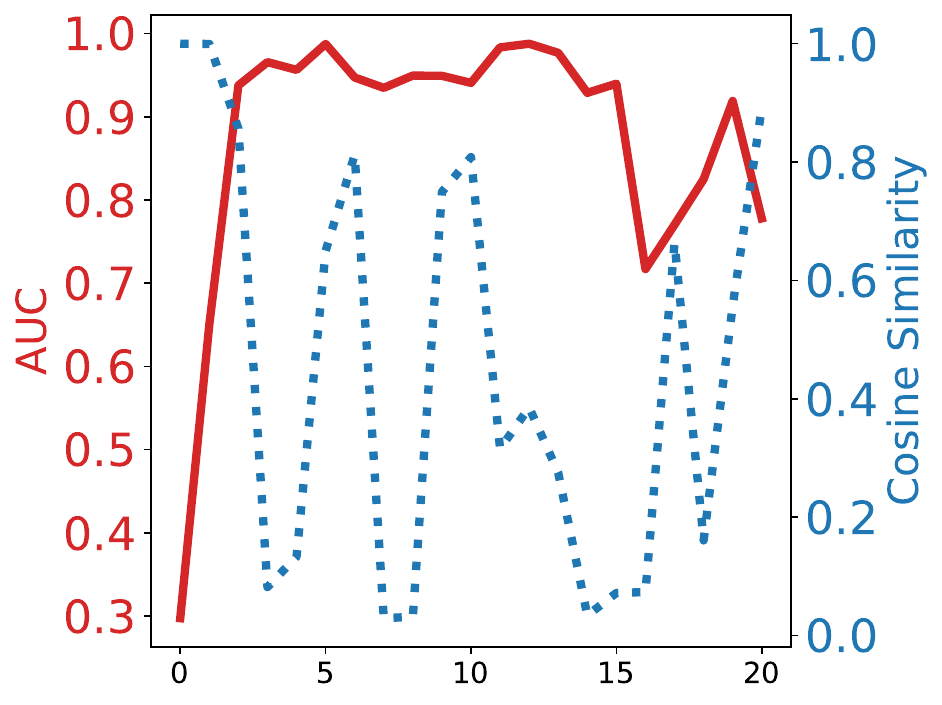}
        \caption{
        Eigenvector estimates \& AUC
        }
        \label{fig:small-NN-aucs-vs-angles}
  \end{subfigure}
  \caption{
  We train a neural network ensemble (Fig. \ref{fig:small-NN-data}).
  We compute underspecification scores (solid line), which correlate with the standard deviation of the ensemble (dotted line) (Fig. \ref{fig:small-NN-scores-vs-stdev}).
  Our OOD performance achieves high AUC (solid line) even though some of our eigenvector estimates have low cosine similarity to ground truth (dotted line) (Fig. \ref{fig:small-NN-aucs-vs-angles}).
  }
  \label{fig:small-NN-lanczos-example}
\end{figure}
\begin{wrapfigure}{R}{0.3\textwidth}
  \centering
  \includegraphics[scale=0.27]{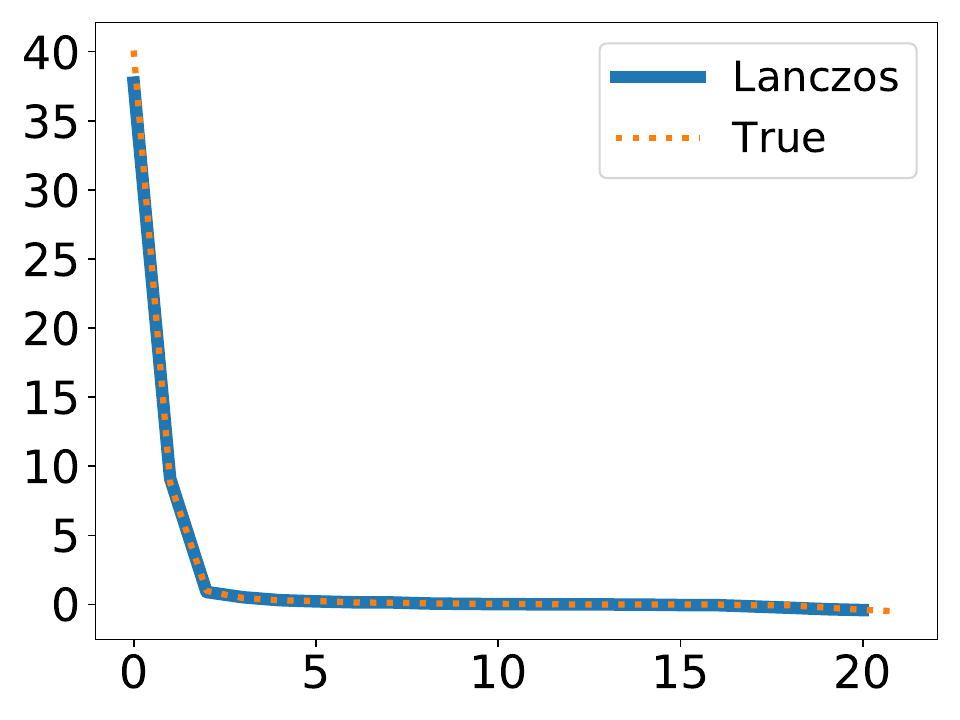}
  \caption{ True and estimated eigenspectrums for toy model Hessian.
  We note that the first few eigenvalues account for most of the variation, and that our estimates are accurate.
  }
  \label{fig:small-nn-eigenvalue-accuracy}
\end{wrapfigure}
We begin with an easily visualized toy experiment.
In Fig. \ref{fig:small-NN-data}, we show our data ($y = \sin{4x} + \mathcal{N}(0, \frac{1}{4})$).
We generate in-distribution training data from $x \in [-1, 0] \cup [1, 2]$, but at test time, we consider all $x \in [-3, 4]$.
We train 20 neural networks with the same architecture.
As shown in Fig. \ref{fig:small-NN-data}, the ensemble disagrees most on $x < -1, x > 2$.
This means that we should most mistrust predictions from this model class on these extreme values, since there are many models within the class that perform equally well on the training data, but differ greatly on those inputs.
We should also mistrust predictions from $x \in [0, 1]$, although detecting this underspecification may be harder since the ensemble agrees more strongly on these points.

For each model in the ensemble, we test our method's performance by AUC on an OOD task: can we flag test points which fall outside the training distribution?
We show that the underspecification score is empirically related to the standard deviation of the ensemble's predictions at the input, which in turn is related to whether the input is OOD (Fig. \ref{fig:small-NN-scores-vs-stdev}).
Examining one model from this ensemble, we observe that by estimating only $m = 2$ eigenvectors, we achieve $> 90 \%$ AUC (Fig. \ref{fig:small-NN-aucs-vs-angles}).
It turns out that $m = 10$ performs best on this task/model.
As we complete more iterations ($m > 10$) we start finding smaller eigenvalues, which are more important to the ensemble subspace and whose eigenvector we do not wish to project out.
We note our AUC improves even with some eigenvector estimates having low cosine similarity to the ground truth eigenvectors (the Lanczos iteration has some stochasticity due to minibatch estimation --- see Appendix \ref{app:lanczos} for details).
We hypothesize this robustness is because the ensemble subspace of this model class is relatively low-dimensional.
Even if an estimated vector is noisy, the non-overlapping parts of the projection will likely be mostly perpendicular to the ensemble subspace, due to the properties of high-dimensional space.

\subsection{Simulated Features} \label{sec:experiments-simulated-features}

\begin{figure}
  \includegraphics[width=.25\textwidth]{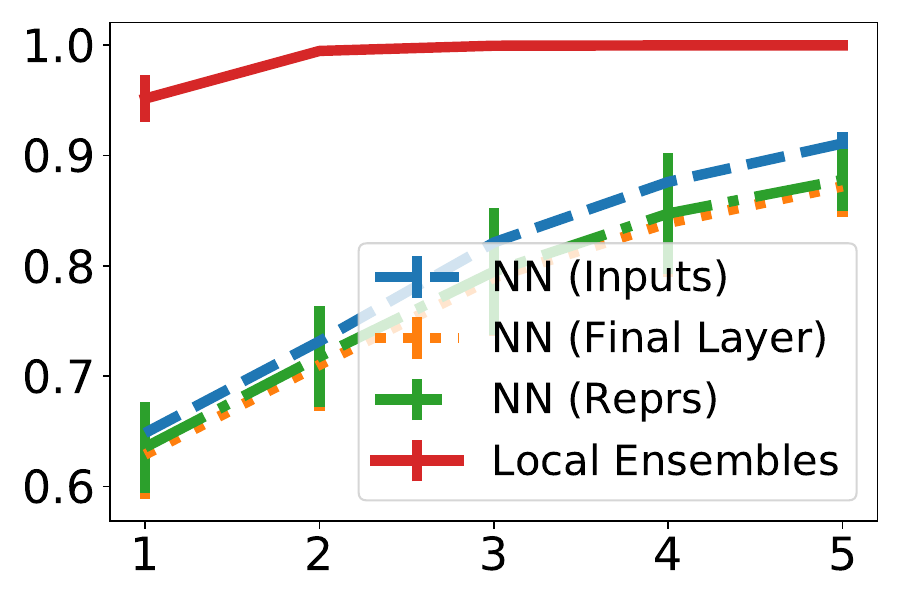}\hfill
  \includegraphics[width=.25\textwidth]{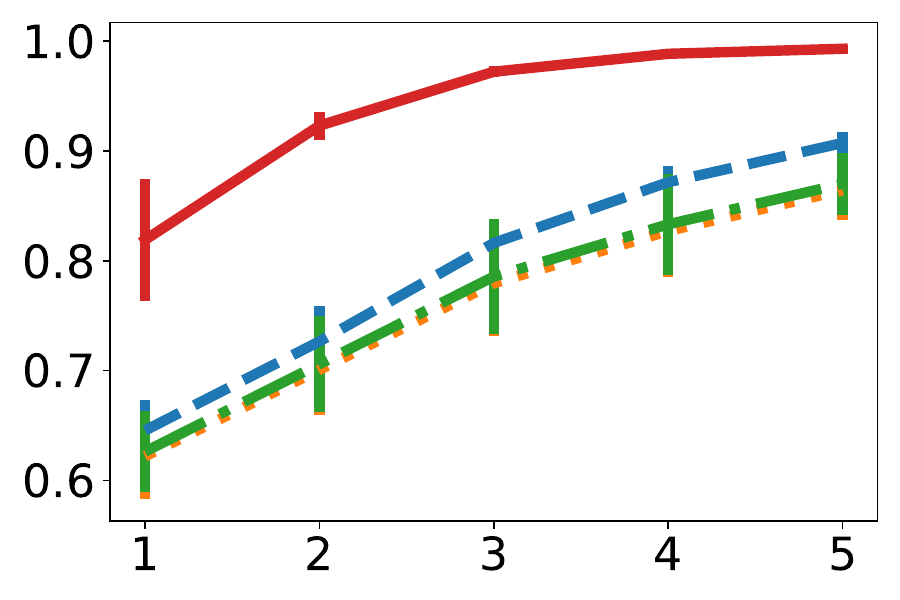}\hfill
  \includegraphics[width=.25\textwidth]{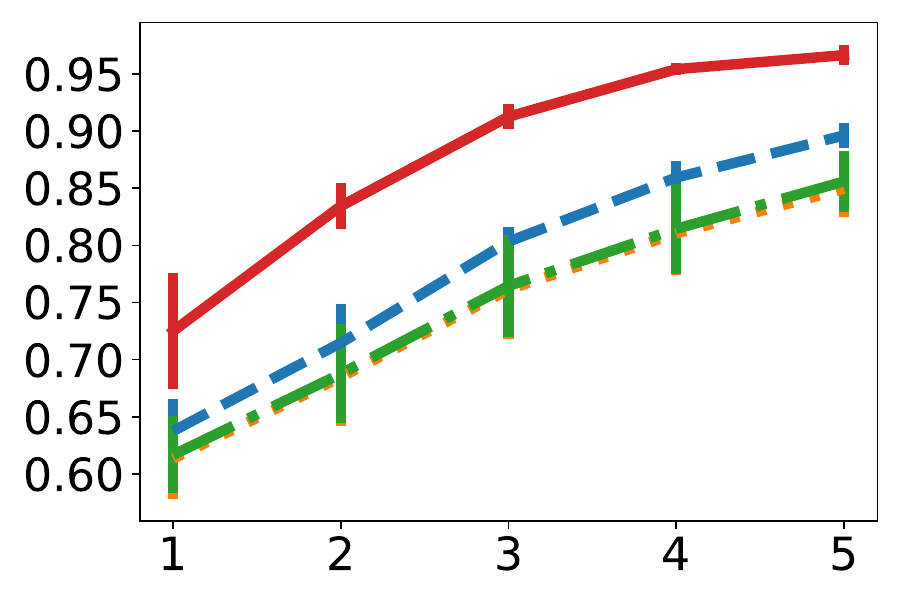}\hfill
  \includegraphics[width=.25\textwidth]{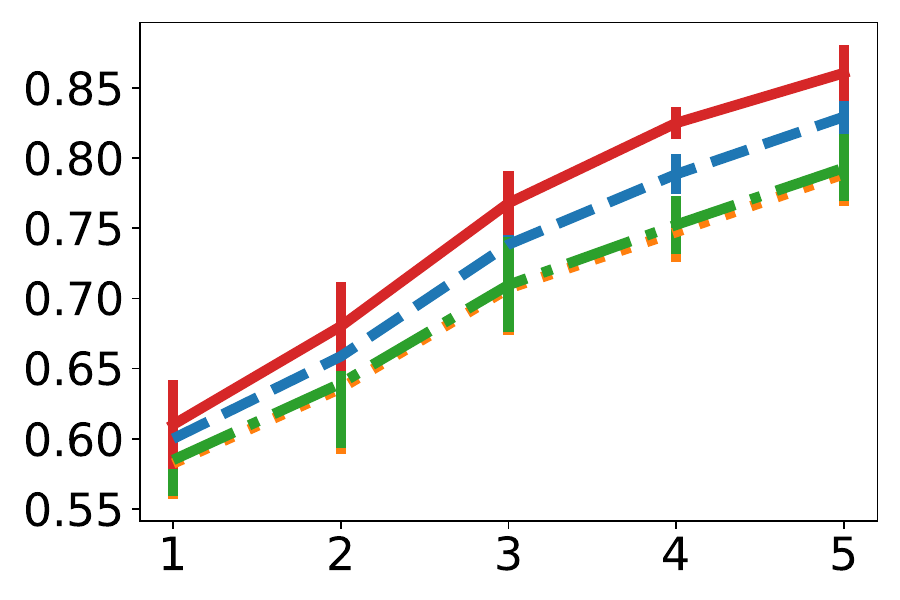}\hfill
  \caption{
  AUC achieved on WineQuality simulated features task (y-axis) compared to the number of extra features simulated (x-axis).
  Noise parameter $\sigma$ increased from left to right $\sigma \in \{0, 0.1, 0.2, 0.5\}$.
  Solid line is our method.
  Results averaged across 5 random seeds, standard deviations shown.
  Results for the other three datasets are qualitatively similar and shown in Appendix \ref{app:simulated-features}.
  }
  \label{fig:simulated-features}
\end{figure}

In this experiment, we create a blind spot in a given dataset by extending and manipulating the data's feature distribution.
We induce a collinearity in feature space by generating new features which are a linear combination of two other randomly selected features in the training data.
This means there is potential for underdetermination: multiple, equally good learnable relationships exist between those features and the target.
However, at test time, we will sometimes sample these simulated features from their \textit{marginal} distribution instead.
This breaks the linear dependence, requiring underspecification (the model is by definition underconstrained), without making the new data trivially out-of-distribution.
We can make this underspecification detection task easier by generating several features this way, or make it harder by adding some noise $\sim \mathcal{N}(0, \sigma^2)$ to these features.

We run this experiment on four tabular datasets.
We compare to three nearest-neighbour baselines, where the metric is the distance (in some space) of the test point to its nearest neighbour by Euclidean distance in an in-distribution validation set.
\textit{NN (Inputs)} uses input space;
\textit{NN (Reprs)} uses hidden representation space, which is formed by concatenating all the activations of the network together (inspired by \citet{papernot2018deep}, who propose a similar method for adversarial robustness);
and \textit{NN (Final Layer)}, uses just the final hidden layer of representations.
We note that since our method is \textit{post-hoc} and can be applied to any twice-differentiable pre-trained model, we do not compare to training-based methods e.g. those producing Bayesian predictive distributions \citep{gal2016dropout,blundell2015weight}.
Our metric is AUC: we aim to assign higher scores to inputs which break the collinearity (where the feature is drawn from the marginal), than those which do not.
In Figure \ref{fig:simulated-features}, we show that local ensembles (LE) outperform the baselines for each number of extra simulated features, and that this performance is fairly robust to added noise.

\subsection{Correlated Latent Factors} \label{sec:experiments-latent-factors}

Here, we extend the experiment from Section \ref{sec:experiments-simulated-features} by inducing a blind spot in \textit{latent} space.
The rationale is similar: if two latent factors are strongly correlated at training time, the model may grow reliant on that correlation, and at test-time may be underdetermined if that correlation is broken.

\begin{wraptable}{R}{0.6\textwidth}
\centering
\begin{tabular}{ccccc}
\hline
\textbf{Method}       & \textbf{M/E}         & \textbf{M/H}         & \textbf{A/E}         & \textbf{A/H}         \\ \hline
MaxProb               & 0.738       & \textbf{0.677}      & 0.461                & 0.433                \\
NN (Pixels)           & 0.561                & 0.550                &  \textbf{0.521}       & 0.547                \\
NN (Reprs)            & 0.584                & 0.578                & \textbf{0.503}      & 0.533                \\
NN (Final Layer)      & 0.589                & 0.517                & 0.480                & 0.497                \\
LE (Loss) & \textbf{0.770} & \textbf{0.684}     & 0.454                & 0.456                \\
LE (Predictions) & 0.364                & 0.544                & \textbf{0.519}       &  \textbf{0.582} \\ \hline
\end{tabular}
\caption{
AUC for Latent Factors OOD detection task.
Column heading denotes in-distribution definitions: labels are $M$ (Male) and $A$ (Attractive); spurious correlates are $E$ (Eyeglasses) and $H$ (Wearing Hat).
Image is in-distribution iff label = spurious correlate.
LE stands for local ensembles.
Each Lanczos iteration uses 3000 eigenvectors.
500 examples from each test set are used.
95\% CI is bolded.
}
\label{table:latent-factors}
\end{wraptable}
We use the CelebA dataset \citep{liu2015deep} of celebrity faces, which annotates every image with 40 latent binary attributes describing the image (e.g. "brown hair").
To induce the blind spot, we choose two attributes: a \textit{label} $L$ and a \textit{spurious correlate} $C$.
We then create a training set where $L$ and $C$ are perfectly correlated: a point is only included in the training set if $L = C$.
We train a convolutional neural network (CNN) as a binary classifier to predict $L$.
Then, we create a test set of held-out data where $P(L = C) = P(L \neq C)$.
The test data where $L \neq C$ is in our model's blind spot; these are the inputs for which we want to output high underspecification scores.
We show in Appendix \ref{app:latent-factors} that the models dramatically fail to classify these inputs ($L \neq C$).
We compare to four baseline underspecification scores: the three nearest-neighbour methods described in Sec. \ref{sec:experiments-simulated-features}, as well as \textit{MaxProb}, where we use $1 - $ the maximum outputted probability of the softmax.
We test two values of $L$ (\textit{Male} and \textit{Attractive}) and two values of $C$ (\textit{Eyeglasses} and \textit{WearingHat}).
We chose these specific values of $L$ because they are difficult to predict and holistic i.e.and not localized to particular areas of image space.

In Table \ref{table:latent-factors}, we present results for each of the four $L, C$ settings, showing both the loss gradient and the prediction gradient variant of local ensembles.
Note that the loss gradient cannot be calculated at test time since we do not have labels available --- instead, we calculate a separate underspecification score using the gradient for the loss with respect to each possible label, and take the minimum.
Our method achieves the best performance on most settings, and is competitive with the best baseline on each.
However, the variation between the tasks is quite noteworthy.
We note two patterns in particular.
Firstly, we note that the performance of \textit{MaxProb} and the loss gradient variant of our method are quite correlated, and we hypothesize this correlation is related to $\nabla_{\hat{Y}} \ell$.
Additionally, observe the effect of increasing $m$ is inconsistent between experiments: we discuss possible relationships to the eigenspectrum of the trained models.
See Appendix \ref{app:latent-factors} for a discussion on these patterns.

\subsection{Active Learning}

\begin{figure}[ht!]
  \begin{subfigure}[b]{.5\textwidth}
    \centering
        \includegraphics[scale=0.45]{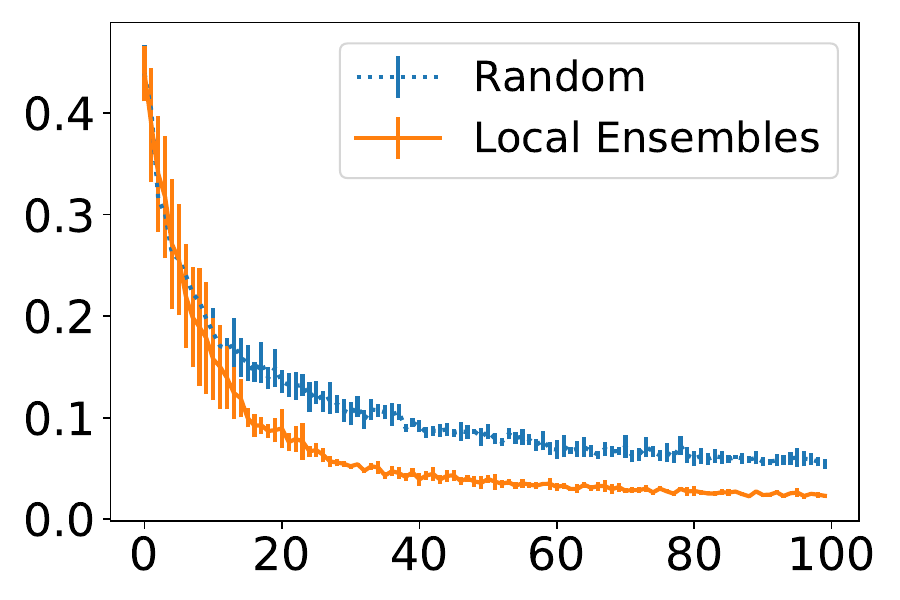}
    	\caption{MNIST}
    	\label{fig:active-learning-MNIST}
  \end{subfigure}
  \begin{subfigure}[b]{.5\textwidth}
    \centering
        \includegraphics[scale=0.45]{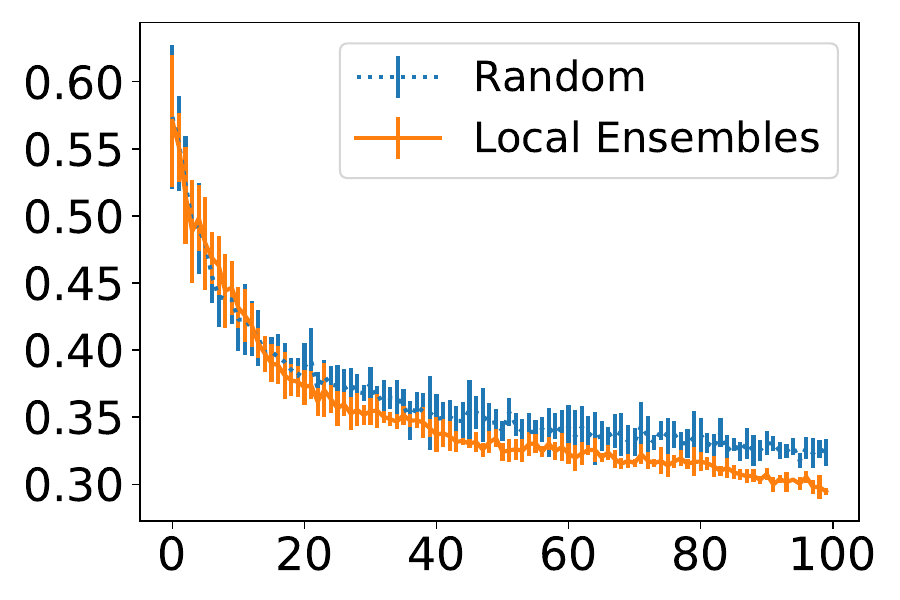}
        \caption{
        FashionMNIST
        }
        \label{fig:active-learning-FashionMNIST}
  \end{subfigure}
  \caption{
  Active learning results. 
  X-axis shows the number of rounds of active learning, Y-axis shows the error rate. 
  Average of 5 random seeds shown with standard deviation error bars.}
  \label{fig:active-learning}
\end{figure}

Finally, we consider the case where we know our model has blind spots, but do not know where they are.
We use active learning to probe this situation,
with the hypothesis that the most useful points to add to our training set may be among the most underspecified.
We use MNIST \citep{lecun2010mnist} and FashionMNIST \citep{xiao2017online} for our active learning experiments.
We begin the first round with a training set of twenty: two labelled data points from each of the ten classes.
In each round, we train to a minimum validation loss using the current training set.
After each round, we select ten new points from a randomly selected pool of 500 unlabelled points, and add those ten points and their labels to our training set.
We compare local ensembles (selecting the points with the highest underspecification scores using the loss-gradient variant) to a random baseline selection mechanism.
In Fig. \ref{fig:active-learning}, we show that our method outperforms the baseline on both datasets, and this improvement increases in later rounds of active learning.
We only used 10 eigenvectors in our Lanczos approximation, which we found to be a surprisingly effective approximation; we did not observe improvement with more eigenvectors.
This experiment serves to emphasize the flexibility of our method: by detecting an underlying property of the model, we are able to use the method for a range of tasks (active learning as well as OOD detection).

\section{Conclusion} \label{sec:conclusion}

We present \textit{local ensembles}, a post-hoc method for detecting underspecification in a trained model.
Our method uses local second-order information to approximate the variance of an ensemble.
We describe how to tractably implement this method using the Lanczos iteration to estimate the largest eigenvectors of the Hessian, and demonstrate its practical flexibility and utility.
Although this method is not a full replacement for ensemble methods, which can characterize more complexity in the loss landscape such as multiple modes, we believe it fills an important role in characterizing one component poor prediction reliability.
In future work, we hope to scale up these methods to larger models and datasets and to further explore the properties of different stopping points $m$.
We also hope to explore applications in fairness and interpretability, where understanding model and training bias is of paramount importance.

\subsubsection*{Acknowledgments}
Thanks to Jamie Smith, Yaniv Ovadia, Yoni Halpern, Pang Wei Koh, Jackson Wang, James Lucas, Marc-Etienne Brunet, Kamyar Ghasemipour and Elliot Creager for helpful comments and discussions.

\bibliography{iclr2020_conference}
\bibliographystyle{iclr2020_conference}


\clearpage
\appendix
\section{Ensembles: Other Datasets} \label{app:other-ensembles}

\begin{figure*}
  \includegraphics[width=.3\textwidth]{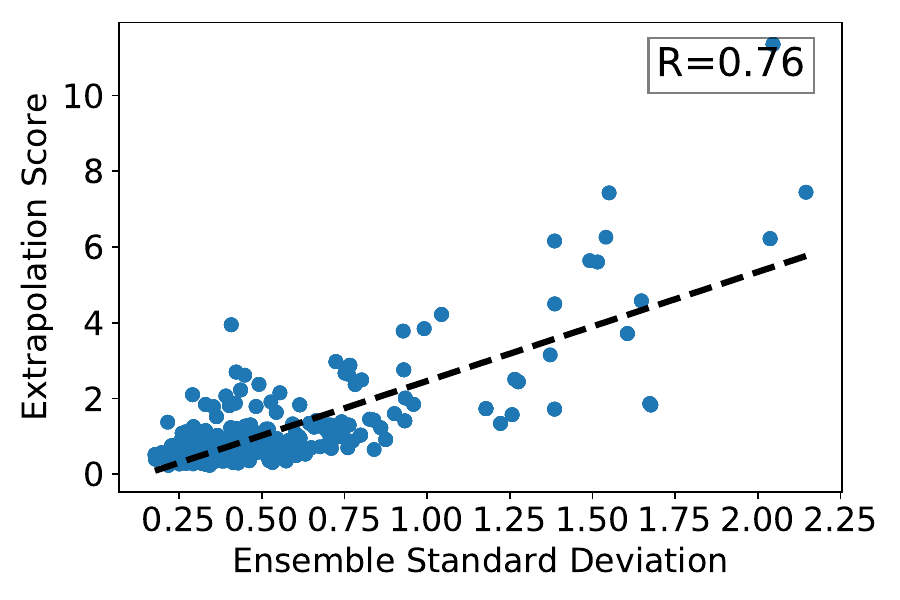}\hfill
  \includegraphics[width=.3\textwidth]{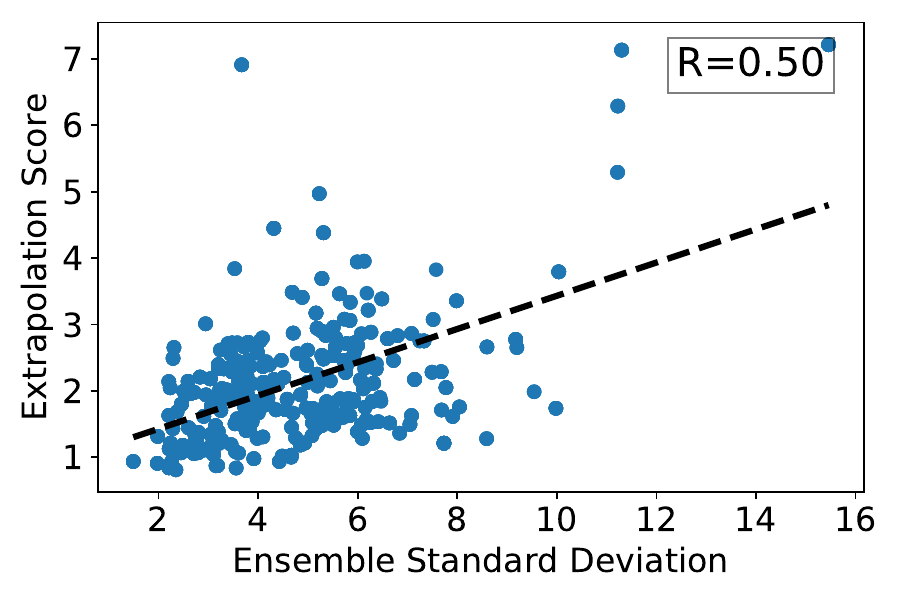}\hfill
  \includegraphics[width=.3\textwidth]{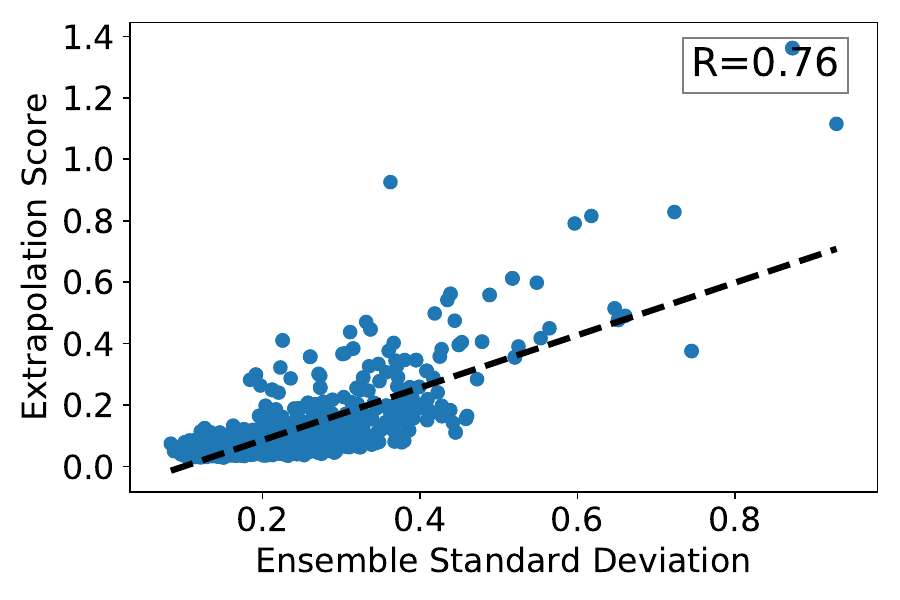}
  \caption{Datasets are (left to right) Boston, Diabetes, Abalone.
  Dotted line represents linear fit of data.
  R is Pearson correlation coefficient.
  }
  \label{fig:other-ensembles}
\end{figure*}

\begin{figure*}
  \includegraphics[width=.3\textwidth]{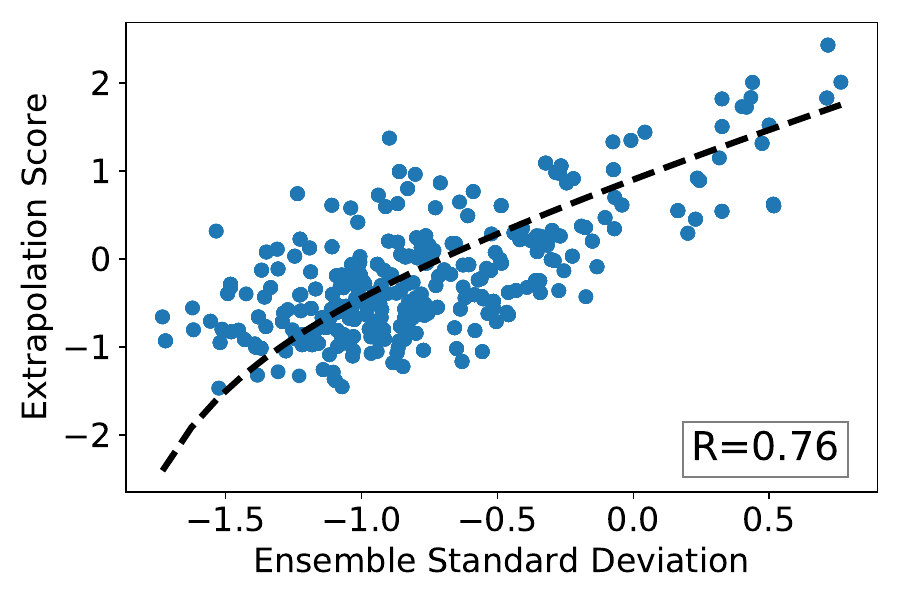}\hfill
  \includegraphics[width=.3\textwidth]{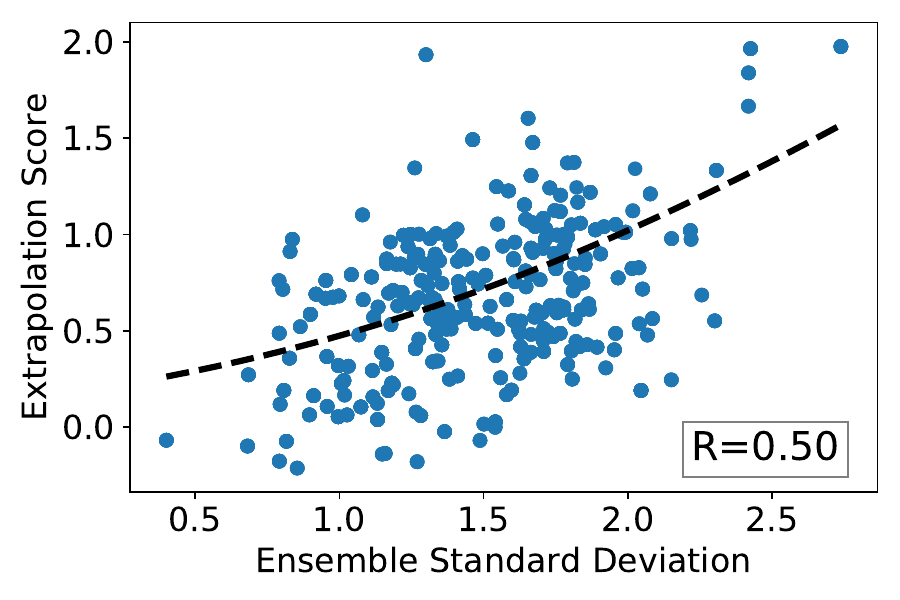}\hfill
  \includegraphics[width=.3\textwidth]{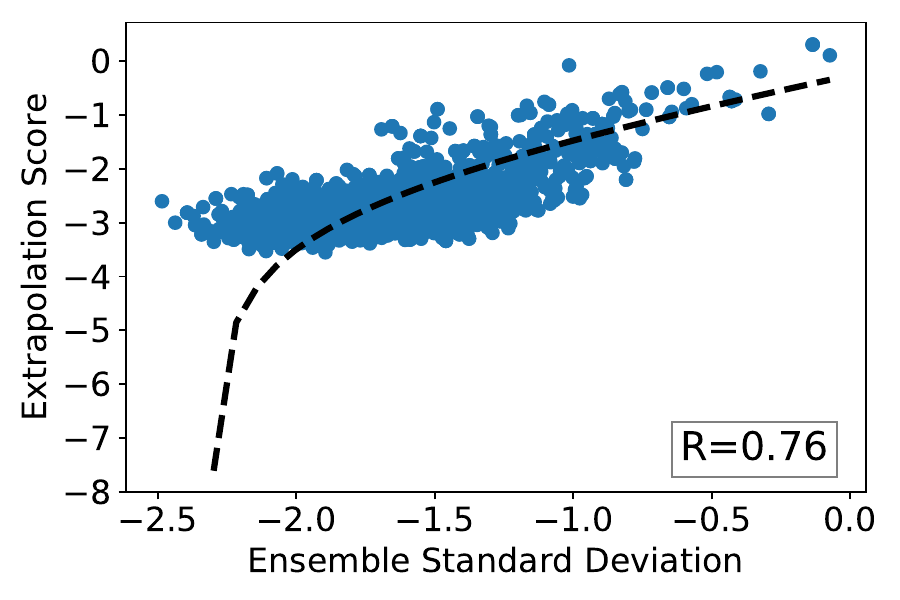}
  \caption{Datasets are (left to right) Boston, Diabetes, Abalone.
  Each axis is log-scaled. 
  Dotted line represents linear fit of data.
  R is Pearson correlation coefficient.
  }
  \label{fig:other-ensembles-log}
\end{figure*}

We show in Figures \ref{fig:other-ensembles} and \ref{fig:other-ensembles-log} the same plots as Fig. \ref{fig:ensembles} for other tabular datasets, demonstrating the strongly linear relationship between underspecification score and ensemble standard deviation.

\section{The Lanczos Iteration: Futher Details} \label{app:lanczos}

The Lanczos iteration \citep{lanczos1950iteration} is a method for tridiagonalizing a Hermitian matrix.
It can be thought of as a variant of power iteration, iteratively building a larger basis through repeatedly multiplying an initial vector by $M$, ensuring orthogonality at each step.
Once $M$ has been tridiagonalized, computing the final eigendecomposition is relatively simple --- a number of specialized algorithms exist which are relatively fast ($O(p^2)$) \citep{dhillon1997new,cuppen1980divide}.

The Lanczos iteration is simple to implement, but presents some challenges.
The first challenge is numerical instability --- when computed in floating point arithmetic, the algorithm is no longer quarnteed to return a good approximation to the true eigenbasis, or even an orthogonal one.
As such, the standard implementation of the Lanczos iteration is unstable, and can be inaccurate even on simple problems.
Fortunately, solutions exist: a procedure known as two-step classical Gram-Schmidt orthogonalization --- which involves ensuring \textit{twice} that each new vector is linearly independent of the previous ones --- is guaranteed to produce an orthogonal basis, with errors on the order of machine roundoff \citep{giraud2005rounding}.
A second potential challenge is presented by the stochasticity of minibatch computation.
Since we access our Hessian $H$ only through Hessian-vector products, we must use only minibatch computation at each stage.
This means that each iteration will be stochastic, which will decrease the accuracy (but not the orthogonality) of the eigenvector estimates provided.
However, in practice, we found that even fairly noisy estimates were nonetheless useful --- see Sec. \ref{sec:experiments-toy} for more discussion.

\subsection{Lanczos Algorithm Code Snippet} \label{app:lanczos-short-code}

The Lanczos algorithm is quite simple to implement.
Figure \ref{fig:lanczos-short-code} shows a a short implementation using Python/Numpy \citep{oliphant2006guide}.

{\tiny
\begin{figure}[ht!]

\begin{verbatim}
import numpy as np

def lanczos(matmul_fn, dim, num_iters, eps=1e-8):
    '''Given implicit access to a matrix M of size (dim x dim) 
    through mamtul_fn, tridiagonalize M into diagonal elements 
    Alpha and off-diagonal elements Beta. Also return the 
    associated orthonormal basis Q.'''

    # Initialize orthonormal basis.
    Q = [np.zeros((dim, 1))]
    # Initialize off-diagonal elements.
    Beta = []
    # Initialize diagonal elements.
    Alpha = []

    # Begin with random initial vector.
    q = np.random.uniform(size=(dim, 1))
    Q.append(q / np.linalg.norm(q))
    Q_k_range = Q[0]

    for k in range(1, num_iters + 1):
        # Compute next step of power iteration.
        z = matmul_fn(Q[k])
        Alpha.append(np.matmul(Q[k].T, z))
        Q_k_range = np.concatenate([Q_k_range, Q[k]], axis=1)

        # Reorthogonalize using two-step classical Gram-Schmidt.
        for _ in range(2):
            z_orth = np.sum(np.matmul(z.T, Q_k_range) * 
                    Q_k_range, axis=1, keepdims=True)
            z = z - z_orth

        Beta.append(np.linalg.norm(z))
        Q.append(z / Beta[k])
        # Check for convergence.
        if np.linalg.norm(z) < eps: 
            break

    return Q, Beta, Alpha

\end{verbatim}
  \caption{
    Example Python implementation of Lanczos algorithm for tridiagonalizing an implicit matrix $M$.
  }
  \label{fig:lanczos-short-code}
  
\end{figure}
}
\section{Simulated Features - Other Datasets} \label{app:simulated-features}

\begin{figure}
  \includegraphics[width=.25\textwidth]{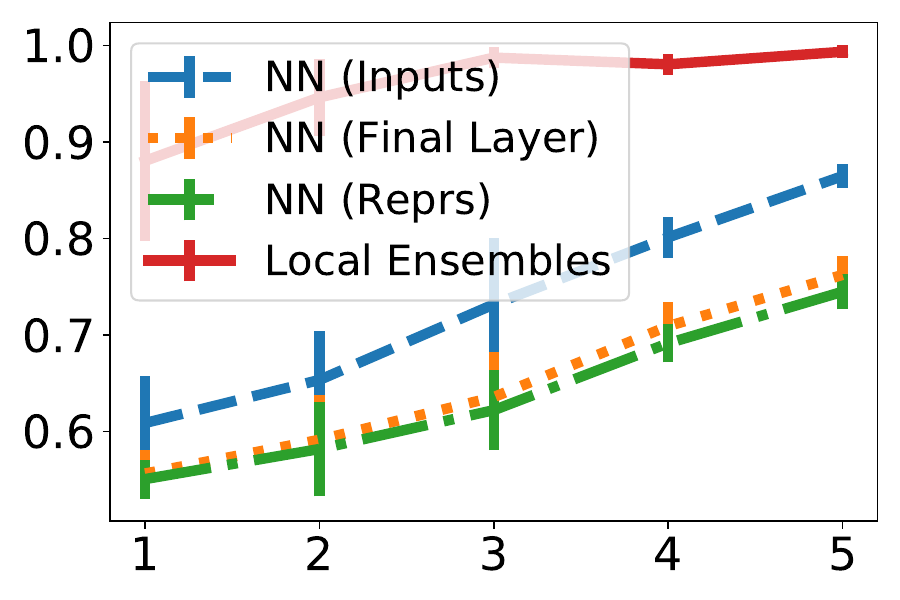}\hfill
  \includegraphics[width=.25\textwidth]{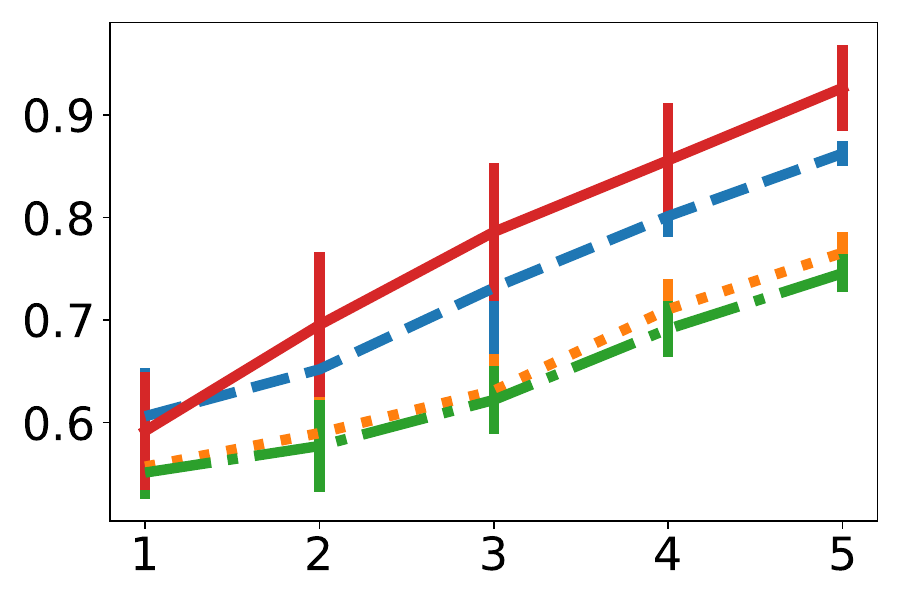}\hfill
  \includegraphics[width=.25\textwidth]{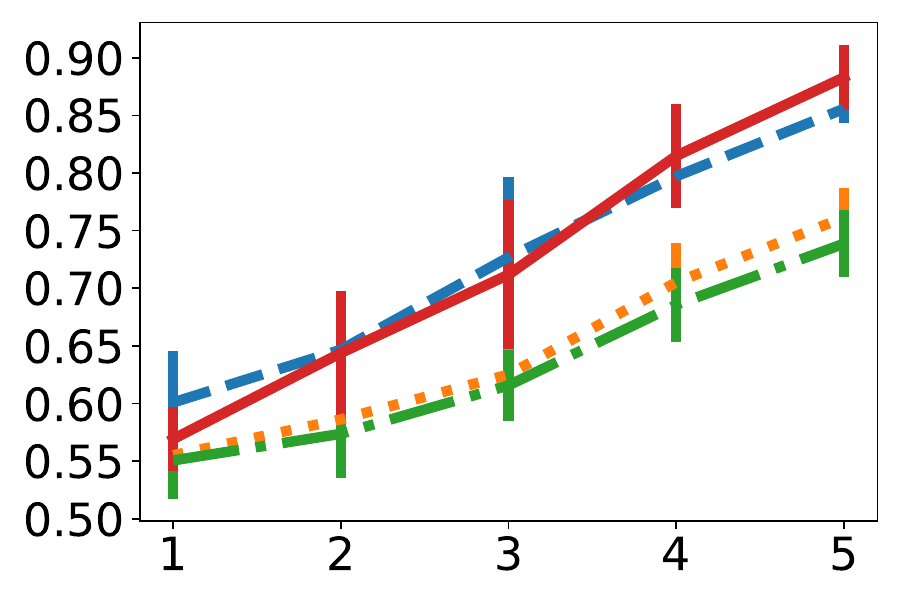}\hfill
  \includegraphics[width=.25\textwidth]{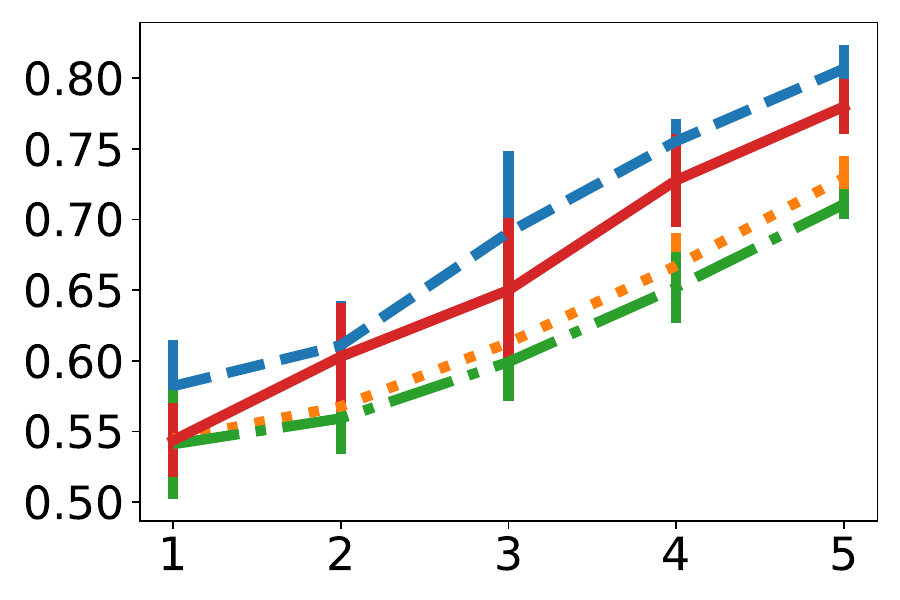}\hfill
  \caption{
  Boston dataset
  }
  \label{fig:simulated-features-boston}
\end{figure}

\begin{figure}
  \includegraphics[width=.25\textwidth]{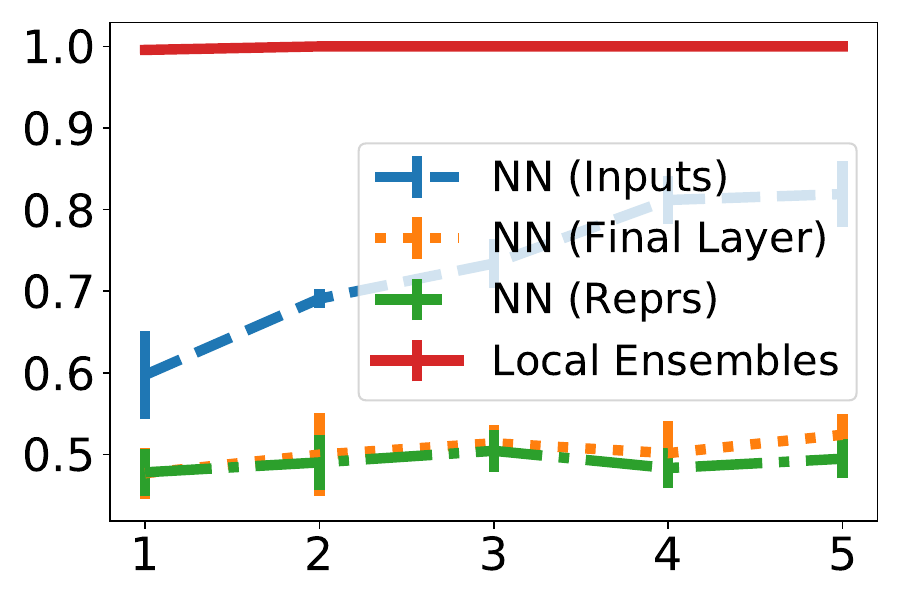}\hfill
  \includegraphics[width=.25\textwidth]{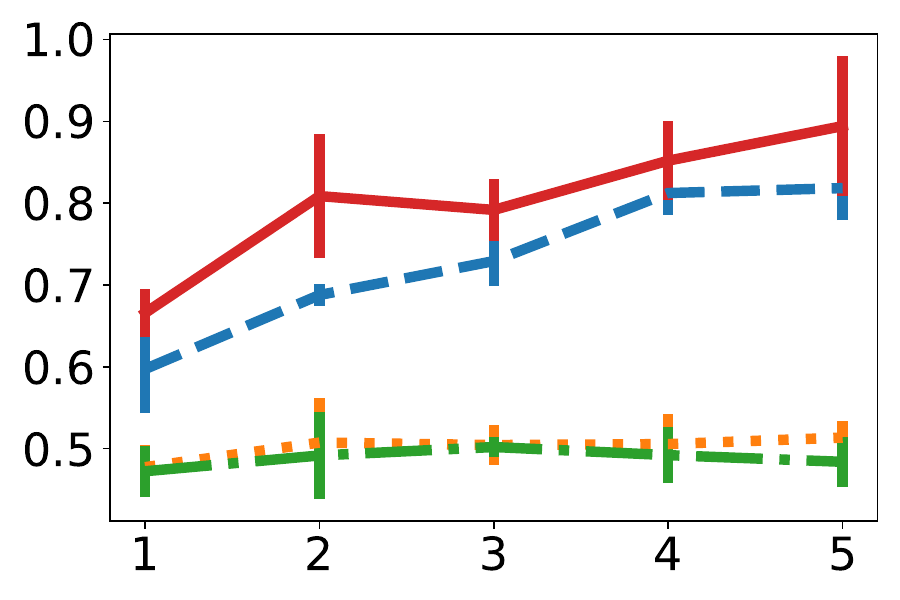}\hfill
  \includegraphics[width=.25\textwidth]{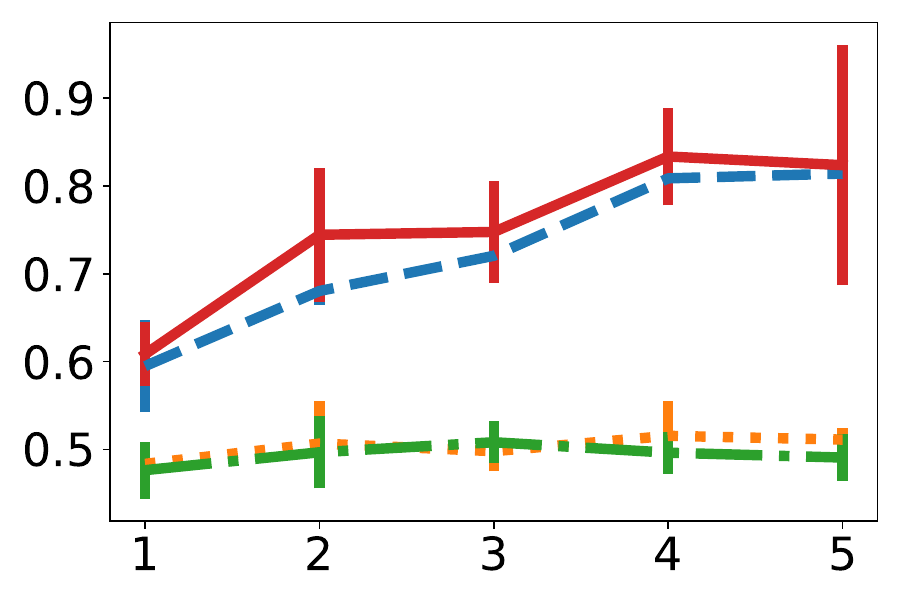}\hfill
  \includegraphics[width=.25\textwidth]{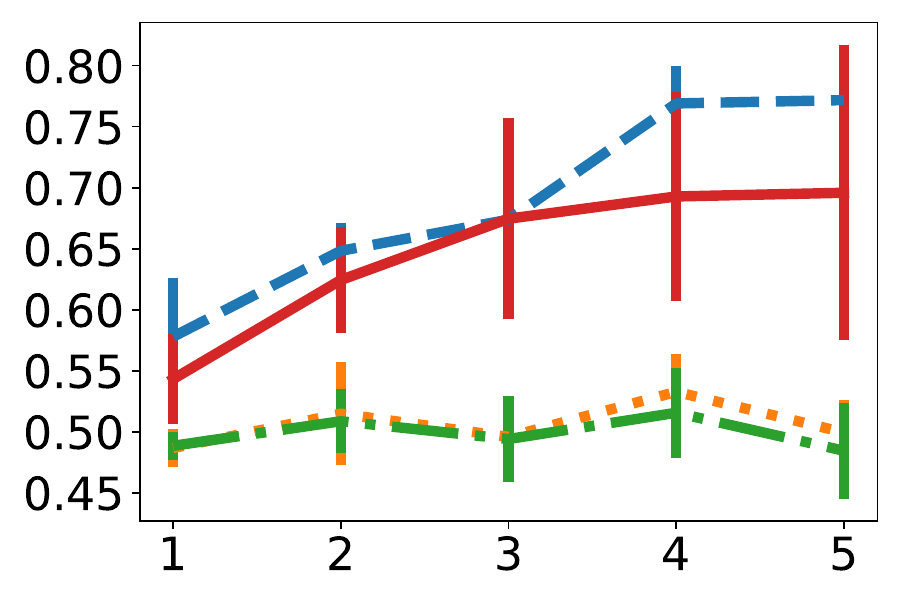}\hfill
  \caption{
  Diabetes dataset
  }
  \label{fig:simulated-features-diabetes}
\end{figure}

\begin{figure}
  \includegraphics[width=.25\textwidth]{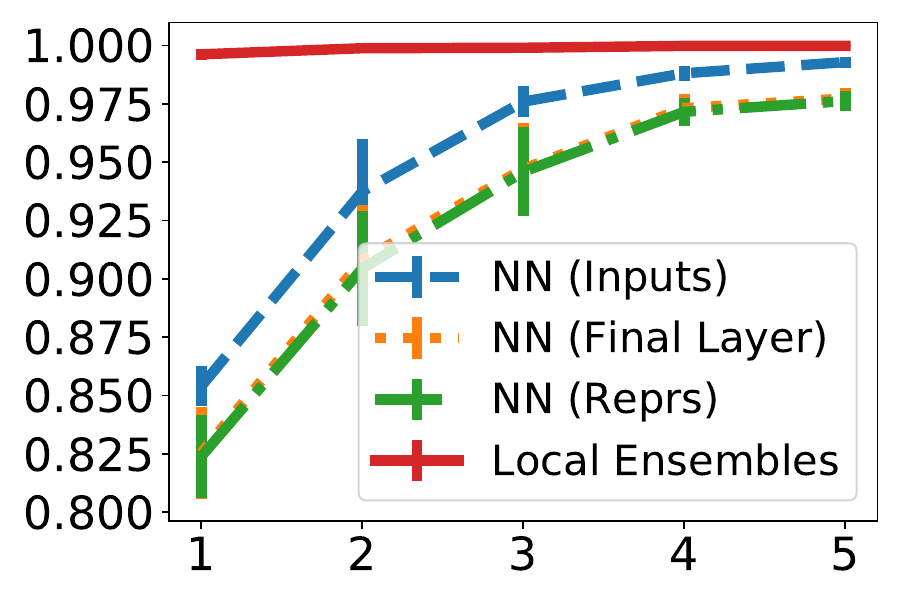}\hfill
  \includegraphics[width=.25\textwidth]{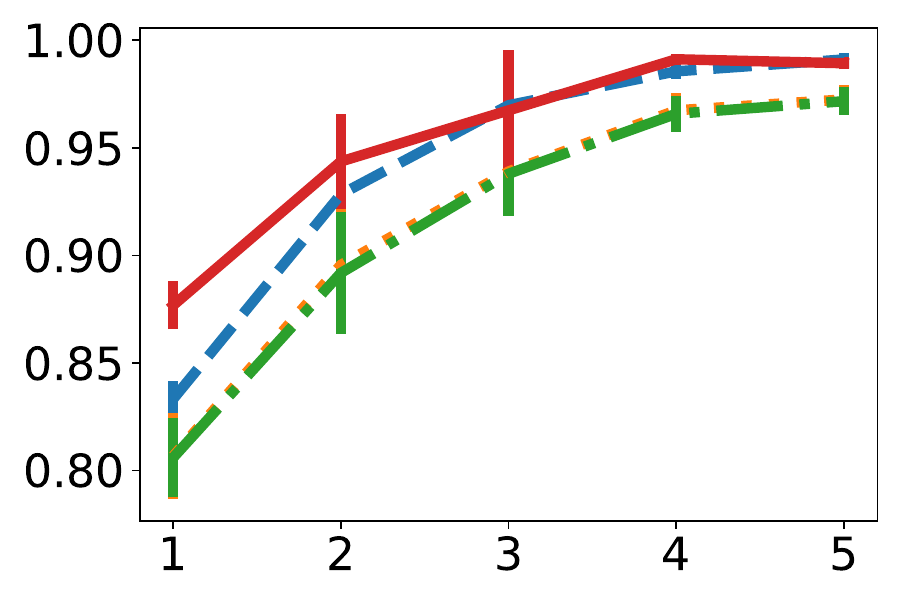}\hfill
  \includegraphics[width=.25\textwidth]{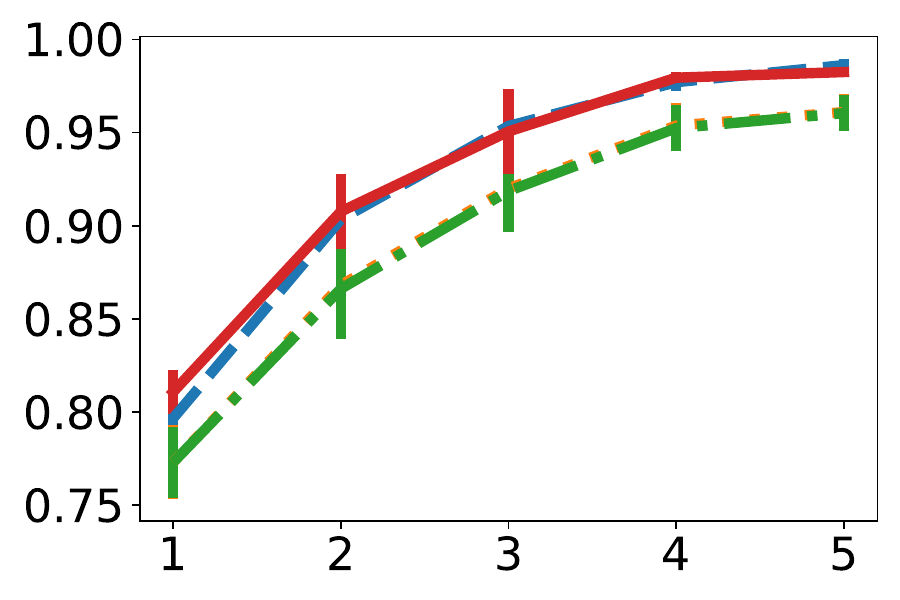}\hfill
  \includegraphics[width=.25\textwidth]{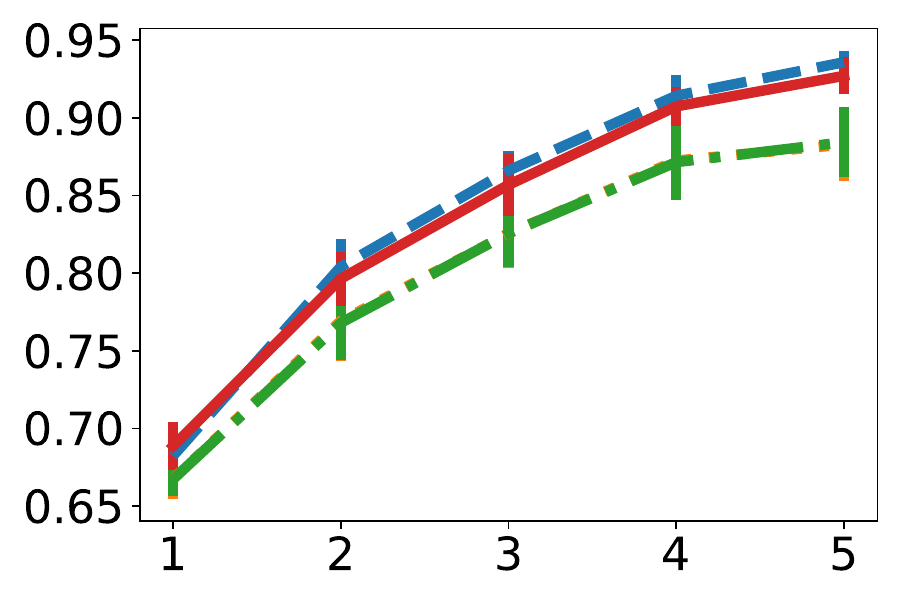}\hfill
  \caption{
  Abalone dataset
  }
  \label{fig:simulated-features-abalone}
\end{figure}

In Section \ref{sec:experiments-simulated-features}, we present results for an experiment where we aim to detect broken collinearities in the feature space.
In Figures \ref{fig:simulated-features-boston}, \ref{fig:simulated-features-diabetes}, and \ref{fig:simulated-features-abalone}, we show results on three more tabular datasets.
See the main text for more experimental details.

\section{Experimental Details} \label{app:data-and-models}

Here we give more experimental details on the datasets and models used.

\subsection{Datasets}

\subsubsection{Tabular Datasets}
We subtracted the mean and divided by the standard deviation for each feature (as calculated from the training set).

\paragraph{Boston \citep{harrison1978hedonic} and Diabetes \citep{efron2004least}.} 
These datasets were loaded from Scikit-Learn \citep{scikit-learn}.

\paragraph{Abalone \citep{nash1994population}.}
This dataset was downloaded from the UCI repository \citep{duaUCI2019} at \url{http://archive.ics.uci.edu/ml/datasets/Abalone}.
We converted sex to a three-dimensional one-hot vector (\textit{M, F, I}).

\paragraph{WineQuality \citep{cortez2009modeling}.}
This dataset was downloaded from the UCI repository \citep{duaUCI2019} at \url{http://archive.ics.uci.edu/ml/datasets/Wine+Quality}.
We used both red and white wines.

\subsection{MNIST, FashionMNIST and CelebA.}
These datasets were loaded using Tensorflow Datasets \url{https://github.com/tensorflow/datasets}.
We divided the pixel values by 255 and, for MNIST and FashionMNIST, binarized by thresholding at 0.7.

\subsection{Experimental Details}

\subsubsection{Toy Data Experiments}
For the first experiment, we train a two-layer neural network with 3 hidden units in each layer and tanh units.
We train for 400 optimization steps using minibatch size 32.
Our data is generated from $y = \sin(4x) + \mathcal{N}(0, \frac{1}{4})$.
We generate 200 training points, 100 test points, and 200 OOD points.
We aggregate $Y$ over a grid of 10 points from -1 to 1, with aggregation function $\min$.
We run the Lanczos algorithm until convergence.

For the second experiment,  we train a two-layer neural network with 5 hidden units in each layer and ReLU units.
Our data is generated from $y = \beta x^2 + \mathcal{N}(0, 1)$.
Our training set consists of $x$ drawn uniformly from $[-0.5, 0.5]$ and $[2.5, 3.5]$.
However, at test time, we will consider $x \in [-3, 6]$.
We generate 200 training points, 100 test points, and 200 OOD points.
We aggregate $Y$ over a grid of 5 points from -6 to 9, with aggregation function $\min$.

\subsubsection{Simulated Features}
For each dataset we use the same setup.
We use a two-layer MLP with ReLU activations and hidden layer sizes of 20 and 100.
We trained all models with mean squared error loss.
We use batch size 64, patience 100 and a 100-step running average window for estimating current performance.
For the Lanczos iteration, we run up to 2000 iterations.
We always report numbers from the final iteration run.
For estimating the Hessian in the HVPs in the Lanczos iteration, we use batch size 32 and sample 5 minibatches.

To pre-process the data, we first split randomly out 30\% of the dataset as OOD.
We choose 2 random features $i, j$ 
and a number $\beta \sim \mathcal{U}(0, 1)$,
and generate the new feature $\tilde{x} = \beta x[i] + (1 - \beta) x[j]$.
We also normalize this feature by its mean and standard deviation as calculated on the training set.
We add random noise to the features after splitting into in-distribution and OOD --- meaning we are not redrawing from the same marginal noise distribution.
We use 1000 examples from in-distribution and OOD for testing.

\subsubsection{Correlated Latent Factors}

We use a CNN with ReLU hidden activations.
We use two convolutional layers with 50 and 20 filters each and stride size 2 for a total of 1.37 million parameters.
We trained all models with cross entropy loss.
We use an extra dense layer on top with 30 units.
We use batch size 32, patient 100 steps, and a 100-step running average window for estimating current performance.
We sample the validation set randomly as 20\% of the training set.
For the Lanczos iteration, we run 3000 iterations.
We always report numbers from the final iteration run.
We use 500 examples from in-distribution and OOD for testing.
For estimating the Hessian in the HVPs in the Lanczos iteration, we use batch size 16 and sample 5 minibatches.

\subsubsection{Active Learning}
We use a CNN with ReLU hidden activations.
We use two convolutional layers with 16 and 32 layers, stride size 5, and a dense layer on top with 64 units.
We trained all models with mean squared error loss.
We use batch size 32, patient 100 steps, and a 100-step running average window for estimating current performance.
\section{Correlated Latent Factors} \label{app:latent-factors}

\subsection{Performance of Binary Classifiers}
In Section \ref{sec:experiments-latent-factors}, we discuss an experiment where we correlated a latent label $L$ and confounder $C$ attribute.
Table \ref{table:latent-factors-accuracy} shows the in-distribution and out-of-distribution test error.
These are drastically different, meaning that learning to detect this type of underspecification is critical to maintain model performance.

\begin{table}[]
\centering
\begin{tabular}{ccccc}
\hline
\textbf{Test Set}   & \textbf{M/E} & \textbf{M/H} & \textbf{A/E} & \textbf{A/H} \\ \hline
In-Distribution     & 0.03         & 0.06         & 0.01         & 0.02         \\
Out-of-Distribution & 0.98         & 0.96         & 0.90         & 0.93         \\ \hline
\end{tabular}
\caption{
Error rate for in and out of distribution test set with correlated latent factors setup.
Column heading denotes in-distribution definitions: labels are $M$ (Male) and $A$ (Attractive); spurious correlate are $E$ (Eyeglasses) and $H$ (Wearing Hat).
Image is in-distribution iff label == spurious correlate.
}
\label{table:latent-factors-accuracy}
\end{table}

\subsection{Behaviour of AUC with More Estimated Eigenvectors}
\begin{figure}
  \includegraphics[width=.25\textwidth]{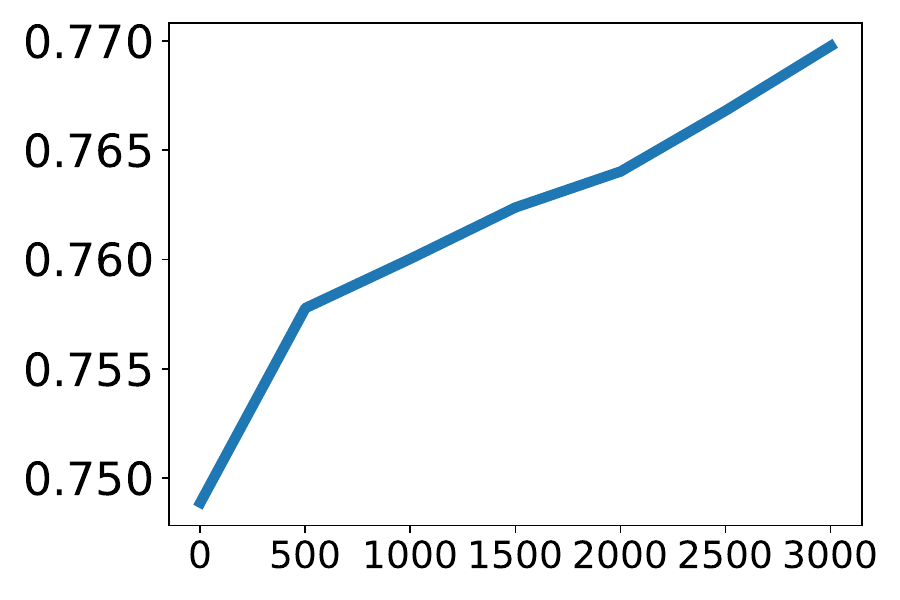}\hfill
  \includegraphics[width=.25\textwidth]{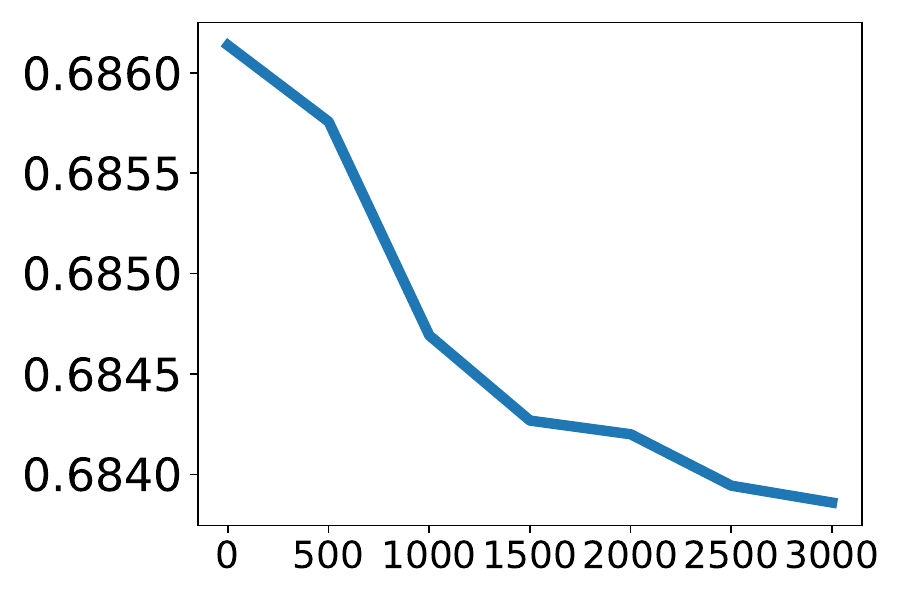}\hfill
  \includegraphics[width=.25\textwidth]{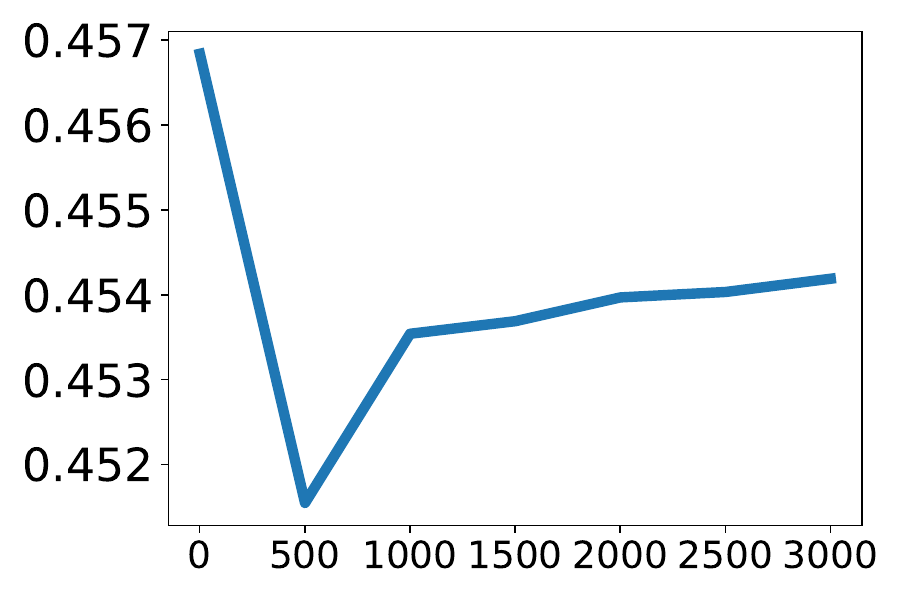}\hfill
  \includegraphics[width=.25\textwidth]{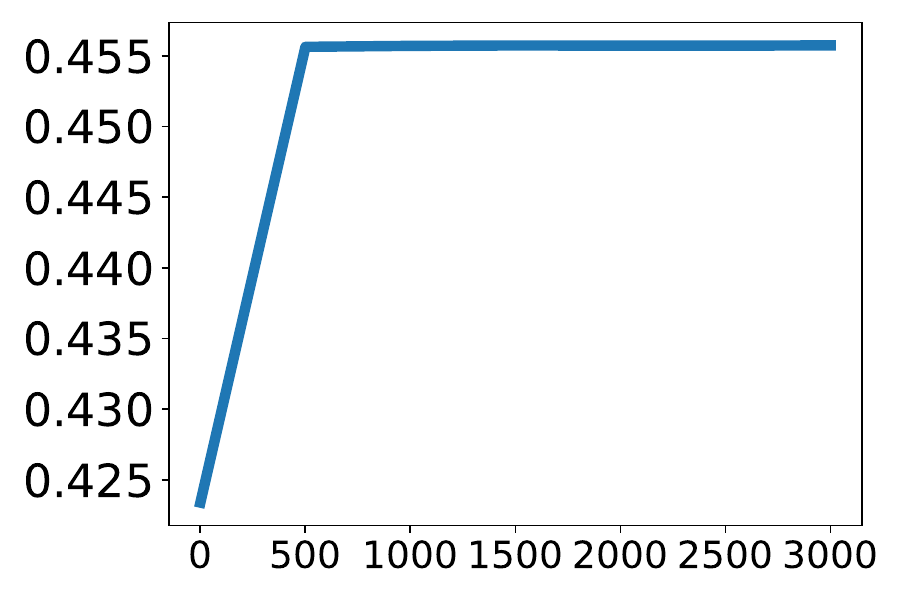}\hfill
  \caption{
  Latent Factors OOD detection task, using loss gradient.
  Y-axis shows AUC, X-axis shows the number of eigenvectors estimated by the Lanczos algorithm, data sampled every 500 eigenvectors.
  Tasks from left to right are \textit{M/E}, \textit{M/H}, \textit{A/E}, \textit{A/H}.
  }
  \label{fig:latent-factors-3000-loss}
\end{figure}

\begin{figure}
  \includegraphics[width=.25\textwidth]{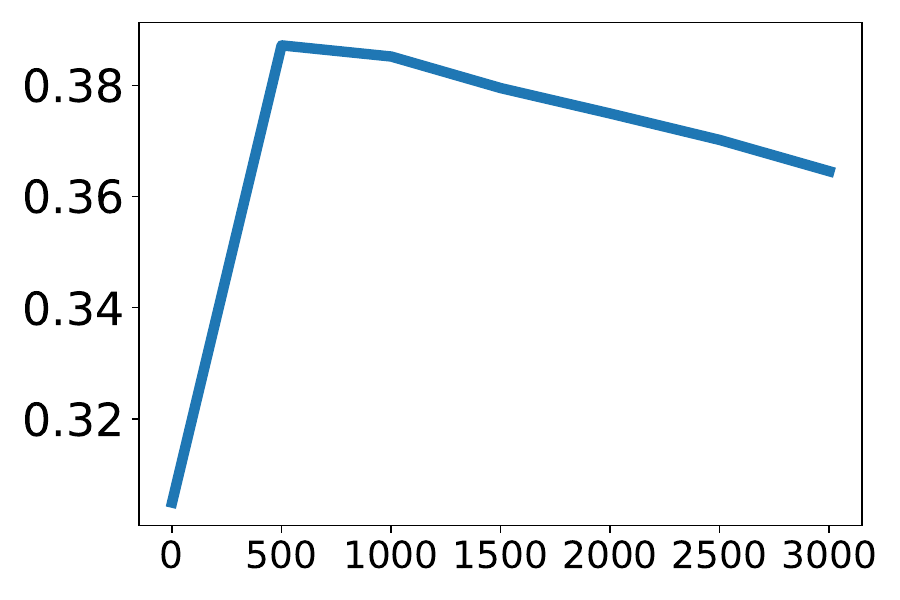}\hfill
  \includegraphics[width=.25\textwidth]{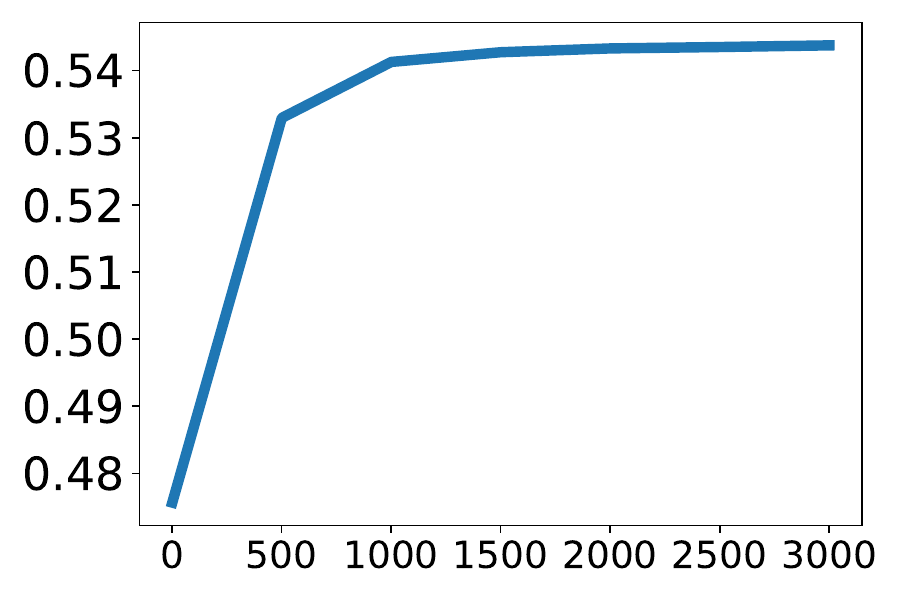}\hfill
  \includegraphics[width=.25\textwidth]{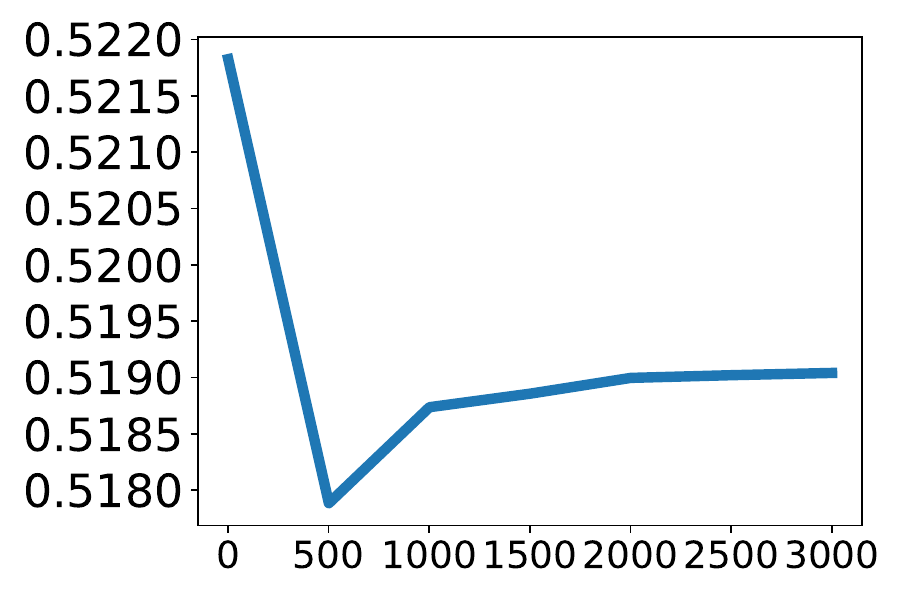}\hfill
  \includegraphics[width=.25\textwidth]{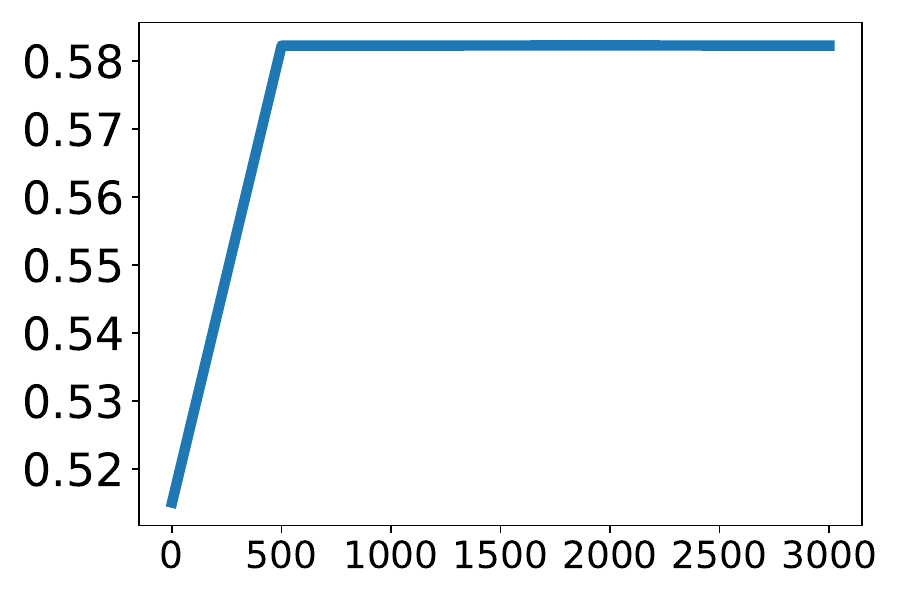}\hfill
  \caption{
  Latent Factors OOD detection task, using loss gradient.
  Y-axis shows AUC, X-axis shows the number of eigenvectors estimated by the Lanczos algorithm, data sampled every 500 eigenvectors.
  Tasks from left to right are \textit{M/E}, \textit{M/H}, \textit{A/E}, \textit{A/H}.
  }
  \label{fig:latent-factors-3000-preds}
\end{figure}

In Fig. \ref{fig:latent-factors-3000-loss} and \ref{fig:latent-factors-3000-preds}, we show that the tasks present differing behaviours as more eigenvectors are estimated.
We observe that for the \textit{Male}/\textit{Eyeglasses} and \textit{Attractive}/\textit{WearingHat} tasks, we get improved performance with more eigenvectors, but for the others we do not necessarily see the same improvements.
Interestingly, this upward trend occurs both times that our method achieves a statistically significant improvement over baselines.
It is unclear why this occurs for some settings of the task and not others, but we hypothesize that this is a sign that the method is working more correctly in these settings.

\subsection{Relationship between Loss Gradient and \textit{MaxProb} Method}

\begin{wrapfigure}{R}{0.3\textwidth}
    \centering
    \includegraphics[scale=0.25]{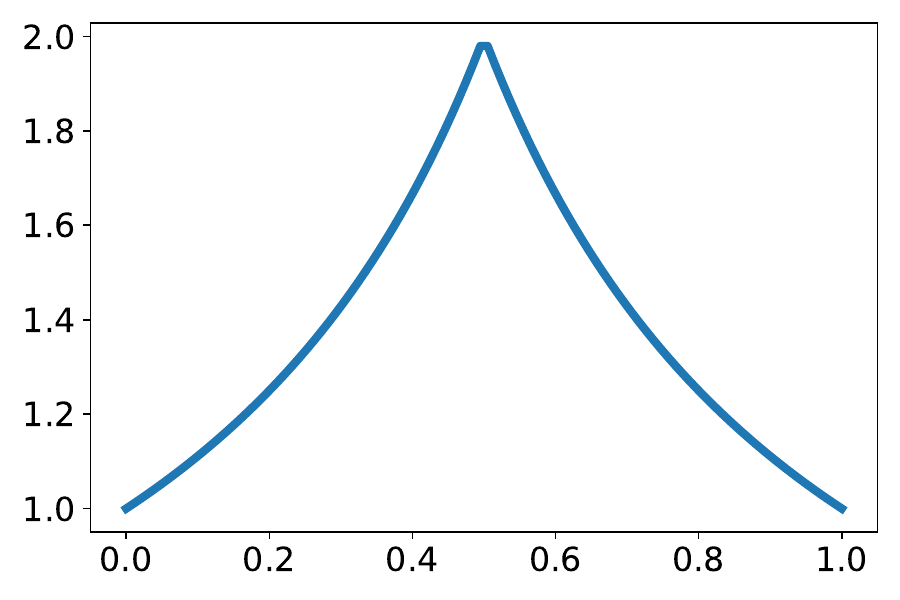}
  \caption{X-axis is $\hat{Y}$, Y-axis is $\min_{Y \in \{0, 1\}} \nabla_{\hat{Y}} \ell(Y, \hat{Y})$.}
  \label{fig:latent-factors-loss-gradient}
\end{wrapfigure}

As discussed in Sec. \ref{sec:experiments-latent-factors}, we have the relationship between the loss $\ell$, prediction $\hat{Y}$, and parameters $\theta$: 
$\nabla_{\theta} \ell = \nabla_{\hat{Y}} \ell \cdot \nabla_{\theta} \hat{Y}$.
Using $\min$ as an aggregation function, we find that  $\min_{Y \in \{0, 1\}} | \nabla_{\hat{Y}} \ell(Y, \hat{Y}) |$ has an inverted V-shape (Fig. \ref{fig:latent-factors-loss-gradient}).
This is a similar shape to $1 - \max(\hat{Y}, 1 - \hat{Y})$, which is the metric implicitly measured by \textit{MaxProb}.

\subsection{Estimated Eigenspectrum of Different Correlated Latent Factor Tasks}

\begin{figure}[ht!]
  \begin{subfigure}[b]{.5\textwidth}
    \centering
        \includegraphics[scale=0.45]{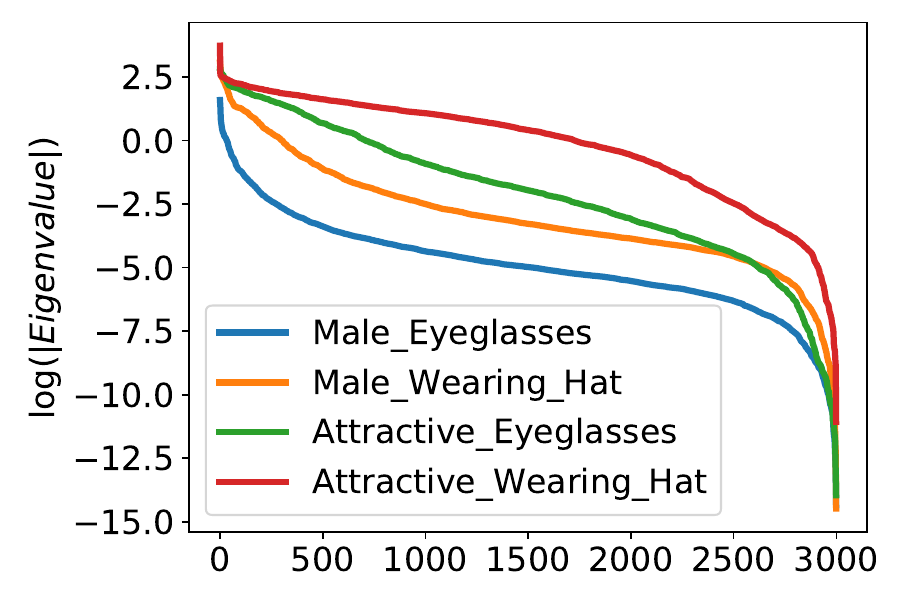}
    	\caption{Log absolute eigenvalues}
    	\label{fig:latent-factors-eigenvalues-log-abs}
  \end{subfigure}
  \begin{subfigure}[b]{.5\textwidth}
    \centering
        \includegraphics[scale=0.45]{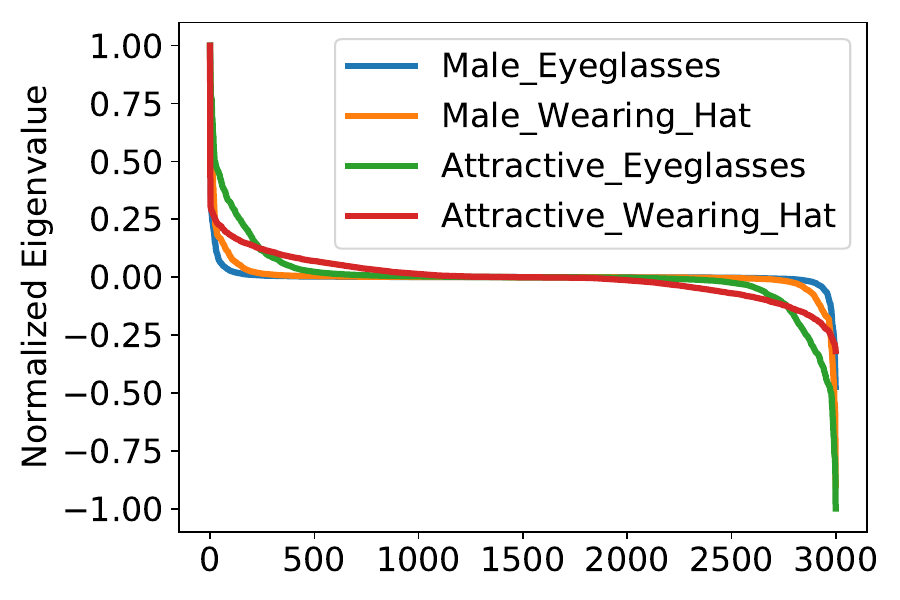}
        \caption{
         Normalized  eigenvalues 
        }
        \label{fig:latent-factors-eigenvalues-norm}
  \end{subfigure}
  \caption{
    We show the estimated eigenspectrum of the four CNNs we train on the correlated latent factors task.
    In Fig. \ref{fig:latent-factors-eigenvalues-log-abs}, we show the absolute estimated eigenvalues sorted by absolute value, on a log scale.
    In Fig. \ref{fig:latent-factors-eigenvalues-norm}, we show the estimated eigenvalues divided by the maximum estimated eigenvalues.
    We only show 3000 estimated eigenvalues because we ran the Lanczos iteration for only 3000 iterations, meaning we did not estimate the rest of the eigenspectrum.
  }
  \label{fig:latent-factors-eigenvalues}
\end{figure}

In Fig. \ref{fig:latent-factors-eigenvalues}, we examine the estimated eigenspectrums of the four tasks we present in the correlated latent factors experiment, to see if we can detect a reason why performance might differ on these tasks.

In Fig. \ref{fig:latent-factors-eigenvalues-log-abs}, we show that the two tasks with the label \textit{Attribute}, the eigenvalues are larger.
These are also the tasks where the loss-gradient-based variant of local ensembles failed, indicating that that variant of the method may be worse at handling larger eigenvalues.

In Fig. \ref{fig:latent-factors-eigenvalues-norm}, we show that the two tasks where the local ensembles methods performed best ($M/E, A/H$, achieving statistically significant improvements over the baselines at a 95\% confidence level, and also showing improvement as more eigenvectors were estimated), the most prominent negative eigenvalue is relatively smaller magnitude compared to the most prominent positive eigenvalue.
This could mean that the local ensembles method was less successful in the other tasks ($M/H, A/E$) simply because those models were not trained close enough to a convex minimum and still had fairly significant eigenvalues.

\end{document}